\documentclass[lettersize,journal]{IEEEtran}
\usepackage{amsmath,amsfonts}
\usepackage{algorithmic}
\usepackage{algorithm}
\usepackage{array}
\usepackage[caption=false,font=normalsize,labelfont=sf,textfont=sf]{subfig}
\usepackage{textcomp}
\usepackage{stfloats}
\usepackage{url}
\usepackage{multirow}
\usepackage{verbatim}
\usepackage{graphicx}
\usepackage{cite}
\usepackage{xcolor}
\usepackage{balance}
\usepackage{soul}
\begin{document}

\title{Graph Lifelong Learning: A Survey}

\author{
Falih Gozi Febrinanto, Feng Xia, Federation University Australia, AUSTRALIA, Kristen Moore, Chandra Thapa, CSIRO's Data61, AUSTRALIA, and Charu Aggarwal, IBM T. J. Watson Research Center, USA

\thanks{Corresponding author: Feng Xia. (e-mail: f.xia@ieee.org)} 
}

\markboth{IEEE Computational Intelligence Magazine, Vol. 00, No. 0, 2022}%
{Febrinanto \MakeLowercase{\textit{et al.}}: Graph Lifelong Learning: A Survey}

\maketitle

\begin{abstract}
Graph learning is a popular approach for performing machine learning on graph-structured data. It has revolutionized the machine learning ability to model graph data to address downstream tasks. Its application is wide due to the availability of graph data ranging from all types of networks to information systems. Most graph learning methods assume that the graph is static and its complete structure is known during training. This limits their applicability since they cannot be applied to problems where the underlying graph grows over time and/or new tasks emerge incrementally. Such applications require a lifelong learning approach that can learn the graph continuously and accommodate new information whilst retaining previously learned knowledge. Lifelong learning methods that enable continuous learning in regular domains like images and text cannot be directly applied to continuously evolving graph data, due to its irregular structure. As a result, \emph{graph lifelong learning} is gaining attention from the research community. This survey paper provides a comprehensive overview of recent advancements in graph lifelong learning, including the categorization of existing methods, and the discussions of potential applications and open research problems.

\end{abstract}


\section{Introduction}
\IEEEPARstart{G}{raph} data is ubiquitous in areas such as social networks, biological networks, road networks, and computer networks \cite{ contribute_wu_GNNSurvey_2019, contribute_zhang_graphSurvey_2020}. Graph learning has shown promising performance in deriving inferences from such graph-structured data~\cite{contribute_feng_graphSurvey_2021}, and in modelling and understanding problems such as drug side effects \cite{intro_drug_zitnik_2018}, knowledge graphs \cite{intro_kg_wang_2018}, social network analysis \cite{intro_socialnetwork_liu_2019}, recommender systems \cite{chen2021heterogeneousRecommender}, and physical movement prediction \cite{kong2022exploringHumanMobility}.

Despite their success, current graph learning approaches are limited in the sense that they can only accommodate predefined tasks~\cite{method_ergnn_zhou_2020, method_continualGNN_wang_2020}. A challenge arises in how to address new tasks that arise sequentially in graph-based problems, such as the addition of new nodes as the graph grows over time and the inclusion of new class labels in classification tasks. Consider, for example, the problem of classifying the topic of each paper in a dataset of scientific publications. Initially, we have a collection of papers covering a fixed set of topics that can be linked to each other via their shared mutual information and thereby represented as a graph. Over time, the number of nodes (which represent papers) and their relations can expand, and new class labels (which represent the topic of the paper) can emerge. When this happens, the model's existing knowledge cannot be used to classify the newly appeared classes because those emerging instances and classes are completely new. A more advanced technique is thus needed to accommodate knowledge from new data and new class labels at different points in time. Another example is traffic flow prediction. Once new locations and new roads appear on a map, a more advanced method is needed to accommodate the new instances and refine the previous experiences to result in good traffic flow prediction in both old and new representations. The focus of this survey paper is a learning model, called \emph{graph lifelong learning}, with the ability to learn new knowledge sequentially in graph-based problems.

\begin{figure}[h]
	\begin{center}
		\includegraphics[scale=0.7]{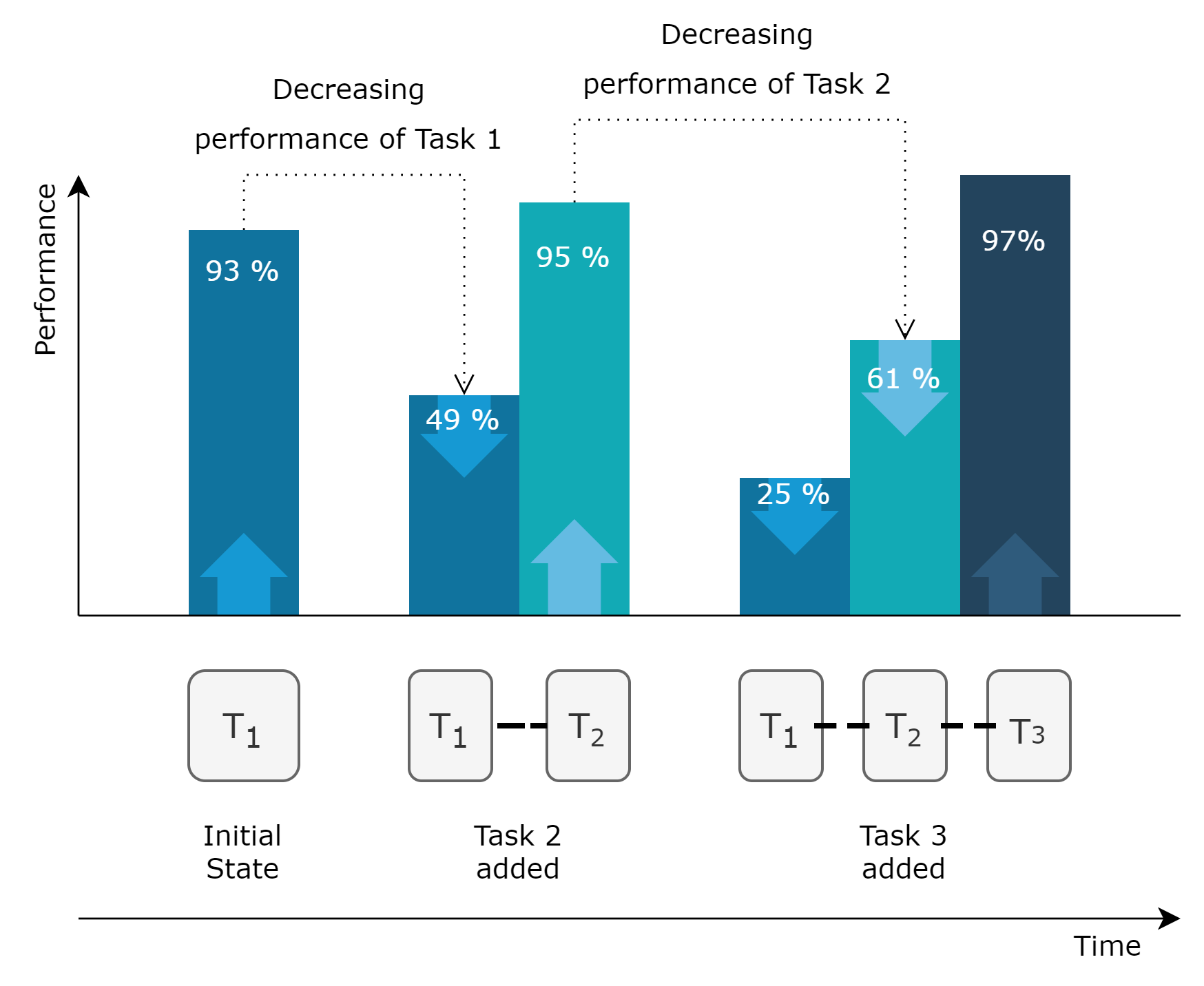}
	\end{center}
	\caption{Illustration of catastrophic forgetting of Task 1 when Tasks 2 and 3 are added.}
	\label{forgetting}
\end{figure}

\begin{figure*}[h]
	\begin{center}
		\includegraphics[scale=0.39]{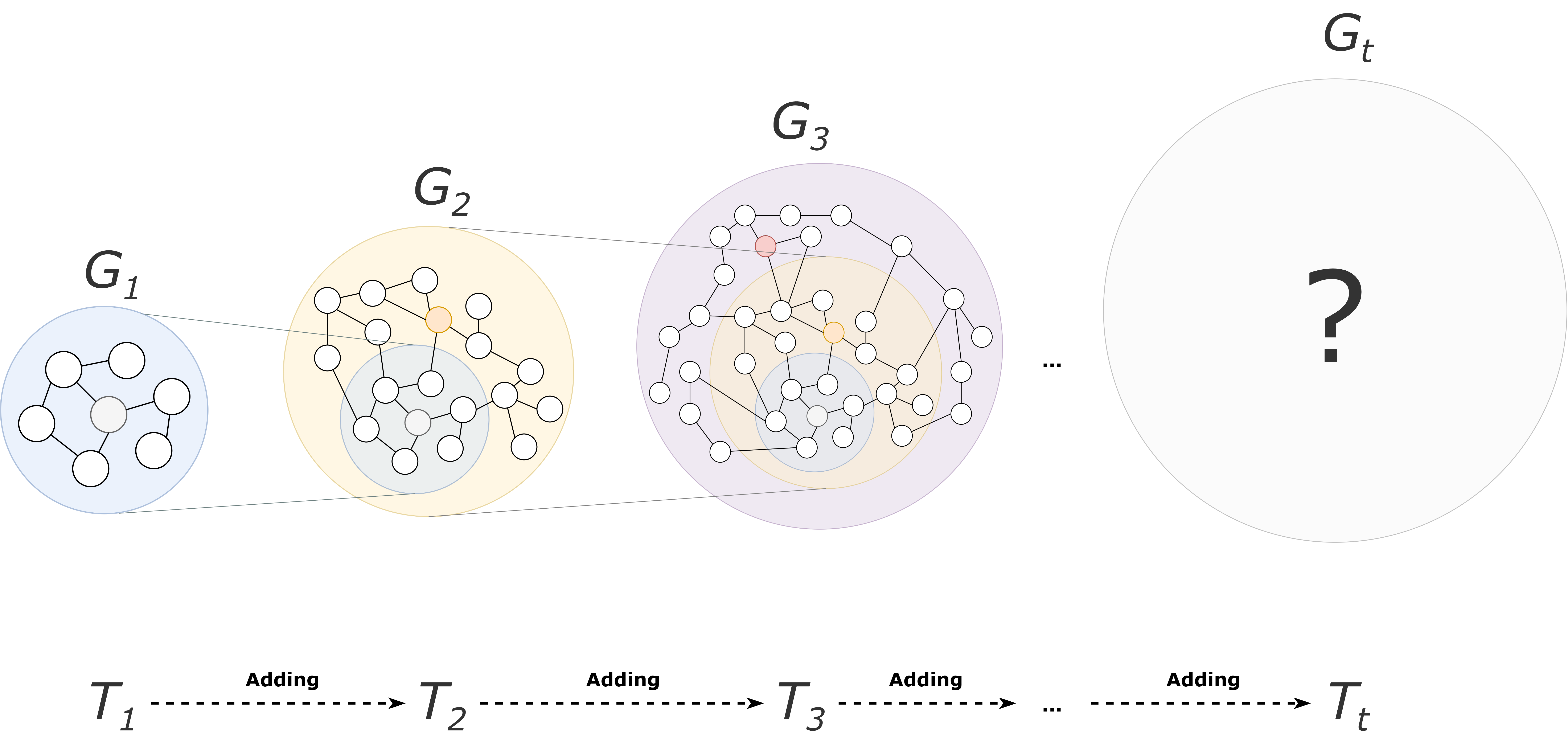}
	\end{center}
	\caption{Graph Evolution. Graph lifelong learning aims to accommodate a new task $T_t$ based on the graph  $G_t = G_{t-1} + \Delta G_t$ while maintaining the performance in solving previous tasks $T_1,T_2,T_3,\dotsc,T_t-_1$.}
	\label{evolving}
\end{figure*}

Lifelong learning, also known as continual learning \cite{lesort_lomonaco_2019_continual_survey}, incremental learning \cite{metrics_first_chaudhry2018riemannian}, and never-ending learning \cite{motive_neverEnding_mitchell}, is a learning characteristic that aims to imitate the human ability to perform continual learning and transfer, acquire, and refine knowledge throughout one's lifetime \cite{chen_liu_2018_LifelongMachineLearning}. A lifelong learning model should be able to recognize the similarity between upcoming and previous knowledge. When a new task is very similar to the prior experience, the learning agent should improve the existing model to perform better. Meanwhile, when a new task has little similarity or correlation with the previous experiences, the system can transfer the knowledge to address the new tasks \cite{motive_definitionLifelong_ruvolo_ELLA}. Most popular deep learning algorithms do not have the ability to incrementally learn incoming new information, and this leads to catastrophic forgetting or interference problems \cite{motive_lifelongDefinition_parisi_2019, motive_catastrophicForgetting_french_1999,motive_catastrophicForgetting_robins_1995}. In particular, new information interferes with previous knowledge, causing a decrease in the model's performance on the old tasks because the new one is overwriting it, as shown in Fig.~\ref{forgetting}. 

Addressing catastrophic forgetting is one of the most important goals of lifelong learning. Prior works have been proposed to enable lifelong learning and address catastrophic forgetting in problems where the input is in the form of a regular structure that has clear order of data, such as images or text data \cite{method_ewc_kirkpatric_2016, method_lwf_li_2016, motive_definitionLifelong_ruvolo_ELLA, method_progNN_rusu_2016_,method_cwr_lomonaco_2017,method_GEM_lopez2017gradient, motive_categorization_maltoni_2019, method_DEN_yoon2017}. However, general lifelong learning approaches used in these domains do not directly generalize to graphs. This is because graphs don't have a unique node ordering or a canonical vector representation, which makes it challenging to define a distance measure between components~\cite{contribute_feng_graphSurvey_2021}. Graph representation learning \cite{openissue_networkEmbedding_cai2018comprehensive} as a canonical framework aims to address this issue by capturing the structure of nodes, edges, and subgraphs and converting them into low-dimensional vectors. While general lifelong learning can accommodate new knowledge added to the model, it does not address the mechanisms of representation learning, which is the main component of graph learning frameworks. There are also challenges in adapting the existing graph representation approaches to lifelong learning. Since conventional approaches such as graph GCNs \cite{motive_GCN_kipf_2016} require the entire graph to perform representation learning, this can introduce efficiency problems when we force such approaches to perform incremental learning. 

In order to successfully apply lifelong learning in the graph domain, existing lifelong learning approaches must be adapted into the specific graph lifelong learning methods that consider the irregular graph structure and use complete components. This requires three main points of uniqueness to be addressed when compared to the traditional lifelong learning:

\begin{itemize}
    \item The ability to detect new instances or tasks in the graph.
    \item The ability to accommodate new knowledge in graphs based on additional data instances and/or classes.
    \item Performing graph representation learning efficiently and accommodating new knowledge without requiring the use of the entire graph structure.
\end{itemize}

\subsection{Motivation of Graph Lifelong Learning}
Graph learning is an advanced method that aims to complete tasks and learn representations on graph-based data. Graph learning can address several problems, such as node prediction\cite{motiveNodeliLearningDeepNeural2019}, edge prediction\cite{motiveLinkcaiLineGraphNeural2021}, and graph prediction\cite{motiveGraphzhangEndtoendDeepLearning2018}. Deep learning in graphs, based on graph neural networks (GNNs), \cite{motiv_gnnModel_franco_2009} refines and extends existing neural network algorithms such as convolutional neural networks (CNNs) and recurrent neural networks (RNNs). Various GNNs models have been proposed, such as graph recurrent neural networks (Graph RNNs) \cite{motive_graphRNN_you_2018}, GCNs \cite{motive_GCN_kipf_2016}, graph autoencoders (GAEs) \cite{motive_GAE_kipf_2016b}, and graph spatial-temporal networks \cite{motive_spatialTemporal_wu_2019}.

Current popular graph learning methods assume that the graph representation is perfect or complete before the learning process begins. However, in many real-world applications, the graph structure can flexibly and dynamically transform when there is a shift in information from neighborhoods over time, as illustrated in Fig. \ref{evolving}. When human social networks grow and evolve, their graph representations accordingly become more complex. That process is based on several factors, such as adding or deleting entities in the graph, changing relationship types, updating entity features, and adding downstream tasks. As with many popular machine learning algorithms, graph learning algorithms are limited by the isolated learning paradigm concept \cite{chen_liu_2018_LifelongMachineLearning}. With that paradigm, the models cannot perform continuous learning to accommodate new tasks and retain past information.

GNNs are able to perform inductive learning\cite{bench_reddit_hamilton2017inductive}, which can accommodate changes to the graph structure. In particular, an inductively trained GNNs can forward-pass newly added nodes/edges, enabling the model to be applied to a different graph without retraining. However, being able to apply a previously trained model to unseen data is not sufficient for incremental learning settings,  as it cannot accommodate cases where new nodes/edges introduce new classes to the classification task. In such a case, the model does not have a knowledge base to classify inputs to the new class~\cite{method_openWorld_2021}, which suggests that another strategy is needed to efficiently retrain the model and accommodate new data and classes for incremental learning.

Addressing the dynamic graph \cite{relatedSurveykazemiRepresentationLearningDynamic2020,relatedSurveybarrosSurveyEmbeddingDynamic2021, overviewDynamicxueDynamicNetworkEmbedding2022} also is not enough for incremental learning. The focus of dynamic graph representation learning is to learn the embedding of the dynamic graphs in every timestamp \cite{dynamic_dyngem_goyalDyngemDeepEmbedding2018,dynamic_dyngraph2vec_GOYAL,dynamic_triad_zhouDynamic2018}. However, since it was not developed with the intention to incrementally solve new tasks, it does not address refining the performance of previous experiences when learning new tasks. In particular,  there is no mechanism to minimize catastrophic forgetting and preserve performance on the old tasks. The difference between relevant graph learning properties is discussed in more detail in Subsection \ref{rlp}. 

The above limitations of graph learning and dynamic graph learning have motivated the development of approaches to extend lifelong learning to graphs in recent years. Graph lifelong learning aims to enable models to acquire new information based on changes to the evolving graph's structure and refinement of previous knowledge in order to handle new unseen tasks on graph-based domains. To achieve this, graph lifelong learning addresses strategies to efficiently retrain the model, decide the right time to retrain the model, and count how much past data should be preserved. In response to the increasing research and development attention on graph lifelong learning, this survey paper comprehensively discusses the current research progress, potential applications, and challenges of graph lifelong learning.

\subsection{Applications}
Graph lifelong learning helps improve learning performance for problems involving dynamic graph data with additional new tasks or new instances. This section discusses various research problems where lifelong learning approaches are needed to mine rich values and solve current and future tasks in graph-related domains.

\paragraph{\textbf{Social networks}}
Graph learning has contributed to solving various prediction problems on social networks \cite{intro_socialnetwork_liu_2019}. The work by Han et al. \cite{applic_continualSocialNet_yi_2020} combines GNNs with a lifelong learning method to detect fake news on social media; however, it does not address the dynamic property of the graph data. Social networks are time-variant by nature, evolving when entities establish or cease relationships. Graph lifelong learning can therefore help adapt the learning process and update knowledge in social network problems like that studied in \cite{applic_continualSocialNet_yi_2020}. 

\paragraph{\textbf{Traffic prediction}}
Traffic prediction is one a critical task in the transportation system that directly affects human daily activities. The challenge is that the underlying network will expand and evolve over time, and the traffic patterns will dynamically adjust to the evolving graph and traffic events that arise. Graph lifelong learning provides excellent potential for predicting streaming traffic flows on the dynamic graph network\cite{method_trafficStream_chen_2021}.

\paragraph{\textbf{Recommender systems}}
Recommender systems are a class of ML algorithms that are designed to recommend items or services to the user based on what the user is most likely to be interested in. Research has shown that recommender systems can be developed using graph learning \cite{appli_recommenderSystem_guo_2019}. In real-world applications, the user interactions and the collection of items or services to be recommended grow and change over time, making it a good candidate for graph lifelong learning.

\paragraph{\textbf{Anomaly detection}}
Anomaly detection is vital in several fields, including finance, security, and environmental monitoring. Graph learning can be used to mine the relationship between entities to capture abnormalities from the system \cite{appli_anomaly_chaudhar_2019}. Most recent approaches ignore temporal information that changes the definitions of an anomaly (e.g., data drift) and the representation of anomaly features in the network over time. This challenge makes graph lifelong learning an interesting candidate to tackle problems, accommodate new knowledge, and prevent the forgetting of previously learned anomaly characteristics.

\subsection{Related Surveys}
Several survey papers discuss the concepts of either lifelong learning or graph learning. The book by Chen and Liu \cite{chen_liu_2018_LifelongMachineLearning} summarised the definition of lifelong learning as a continuous learning process and summarised existing techniques for lifelong learning on different learning approaches such as supervised, unsupervised, semi-unsupervised, and reinforcement learning. Delange et al. \cite{contribute_delange_2021} presented a methodology for dynamically determining the stability-plasticity trade-off in lifelong learning by setting hyperparameters. This survey also discussed the categorization of continual learning methods using a taxonomy illustration. Biesialska et al. \cite{motive_categorization_biesialska_2020_survey} summarized the role of continuous learning in natural language processing. Parisi et al. \cite{motive_lifelongDefinition_parisi_2019} summarised the factors that motivate research in lifelong learning and its relation to the biological aspect. It also explained the lifelong learning setting challenge in neural network models. The above lifelong learning surveys only concentrate on general machine learning or deep learning approaches to address tasks in grid or regular domains but do not cover the concept of lifelong learning in graph data. 

For graph learning, Xia et al. \cite{contribute_feng_graphSurvey_2021} comprehensively reviewed the concept of graph learning with categorizations such as graph signal processing, matrix factorization, random walk, and deep learning, and also explained the application of graph learning in several fields. Wu et al. \cite{contribute_wu_GNNSurvey_2019} provided a comprehensive overview of state-of-the-art GNNs methods and discussed potential directions in different fields. Zhang et al. \cite{contribute_zhang_graphSurvey_2020} reviewed deep learning methods on graphs, such as GCNs, GAEs, graph adversarial methods, graph reinforcement learning (Graph RL), and Graph RNNs.

In addition to those, there are surveys that review other relevant learning in dynamic graphs that mainly focus on graph representation learning and graph embedding in dynamic graphs with different characteristics from lifelong learning. Kazemi et al. \cite{relatedSurveykazemiRepresentationLearningDynamic2020} reviewed representation learning in the dynamic graph, which includes knowledge graphs by categorizing encoders and decoders techniques. It also reviewed potential applications and datasets that are widely used. Barros et al. \cite{relatedSurveybarrosSurveyEmbeddingDynamic2021} provided an overview of dynamic graph embedding, including the underlying methods and the current works so far. Those surveys discuss the embedding process in every timestamp of dynamic graphs that do not consider refining the performance of prior knowledge and performing transfer knowledge to help learn a new task as a lifelong learning characteristic. In Subsection \ref{rlp}, the different traits of other relevant graph learning properties is explained in more detail.

\subsection{Contributions}
Existing surveys in the literature focus only on lifelong learning \textit{or} on graph learning, none address the challenges of uniting the two. This paper provides a survey of recent developments in the emerging field of graph lifelong learning, which enable continuous learning mechanisms in graph learning models. Moreover, this survey paper reviews new perspectives of incremental learning in graph data different from other relevant learning concepts, such as spatio-temporal neural networks and dynamic graph learning. We summarise the benefit of graph lifelong learning that can overcome the catastrophic forgetting problem, refine previous experiences, and accommodate new knowledge to the graph learning model. Based on our knowledge, we are the first to review the emerging topic of graph lifelong learning. Overall, the main contributions of this survey paper are:
\begin{itemize}
  \item A discussion of the potential applications that benefit from implementing graph lifelong learning.
  \item A comprehensive review and categorization of the current progress in graph lifelong learning algorithms.
  \item A comparison of existing models of graph lifelong learning based on its different applications and scenarios.
  \item A discussion of open issues and challenges of graph lifelong learning as insights for future research directions.
\end{itemize}

\subsection{Paper Organization}
The remainder of this survey is organized into the following sections. Section \ref{sec:over} provides an overview of the domain of graph lifelong learning, including lifelong learning in general, relevant graph learning properties, graph lifelong learning scenarios, notation of domain, and taxonomy of the current works. The categorization of graph lifelong learning techniques is further explained in Subsection \ref{sec:taxo}. That categorization includes architectural approaches in Section \ref{arch}, rehearsal approaches in Section \ref{rehear}, regularization approaches in Section \ref{regu}, and hybrid approaches in Section \ref{hybrid}. Section \ref{'sec:method_compare'} compares each categorization to consider its implementation and summarises each method's characteristics based on scenarios and their applications. Section \ref{sec:benchmarks} reviews benchmark settings for graph lifelong learning, including datasets and evaluation metrics. Section \ref{sec:issue} describes the open issues and challenges for future research direction. Lastly, the conclusion of this survey paper is given in Section~\ref{sec:conclu}.

\section{Overview}
\label{sec:over}
This section begins with a discussion of lifelong learning for regular deep learning or machine learning methods in Subsection \ref{sec:life}. Then Subsection \ref{subsection:ML} discusses some related learning paradigms and learning objectives in typical machine learning models and highlights how they differ from lifelong learning. Subsection \ref{rlp} reviews relevant graph learning properties that have similarities with graph lifelong learning problems. Subsection \ref{secenarios} then explains several graph lifelong learning scenarios. Subsection \ref{sec:notationDomain} then explains the notation and problem definition of graph lifelong learning. Finally, Subsection \ref{sec:taxo} presents a categorization of graph lifelong learning approaches that we will review in the later sections.

\subsection{Lifelong Learning}
\label{sec:life}

Lifelong learning aims to enable continuous learning to accommodate newly appeared knowledge and refine previous knowledge. Prior studies on lifelong learning models can be divided into three categories: architectural, regularization, and rehearsal \cite{motive_categorization_maltoni_2019, motive_categorization_biesialska_2020_survey}. 

The architectural approach modifies network architecture by adding more units and expanding and compressing the architecture \cite{motive_definitionLifelong_ruvolo_ELLA, method_cwr_lomonaco_2017, hung2019compacting_CPG}. Rusu et al. \cite{method_progNN_rusu_2016_} introduced a progressive network (ProgNN) that was able to adjust new capacity alongside pre-trained models so that it gave the flexibility to accommodate the latest knowledge and reuse the old ones. Similarly, Yoon et al. \cite{method_DEN_yoon2017} proposed a dynamic expandable network (DEN) that is able to decide the network capacity with only a necessary number of units when new knowledge arrives. 

The regularization approach uses some additional loss terms to maintain the stability of prior parameters when learning new parameters. The most common technique, elastic weight consolidation (EWC), was proposed by Kirkpatrick et al. \cite{method_ewc_kirkpatric_2016}, which regularizes model parameters by penalizing the changes based on task importance, so new tasks will have optimum parameter region where the model performs without any catastrophic forgetting of previous experiences. Li and Hoiem \cite{method_lwf_li_2016} developed a model called learning without forgetting (LWF) that employs distillation loss to maintain the network's output as close to its previous values.

The rehearsal approach relies on a retraining process from prior task samples that can keep the performance of previous experiences \cite{method_exstream_hayes_2019}. Rebuffi et al. \cite{method_icarl_rebuffi_2017} proposed incremental classifier and representation learning (ICARL) that can store samples of each task that is constantly replayed while learning new tasks.

In order to apply lifelong learning models like those mentioned above to problems in graph-based domains, they must first be adapted to incorporate graph learning techniques that address graph representation learning and graph data architecture. This survey paper focuses on reviewing methods that have the specific aim of deploying lifelong learning settings in graph-based domains.
\begin{table*}[]
\tiny
\centering
\caption{Graph Learning Concept Comparison}
\label{conceptCompare}
\resizebox{\textwidth}{!}{%
\begin{tabular}{|l|cccc|}
\hline
\multicolumn{1}{|c|}{\textbf{Properties}} & \multicolumn{4}{c|}{\textbf{Concepts}} \\ \cline{2-5}
\multicolumn{1}{|c|}{} & \multicolumn{1}{c|}{\textbf{\begin{tabular}[c]{@{}c@{}}Graph \\ Neural Network\end{tabular}}} & \multicolumn{1}{c|}{\textbf{\begin{tabular}[c]{@{}c@{}}Graph Spatio \\ Temporal Networks\end{tabular}}} & \multicolumn{1}{c|}{\textbf{\begin{tabular}[c]{@{}c@{}}Dynamic Graph \\ Learning\end{tabular}}} & \textbf{\begin{tabular}[c]{@{}c@{}}Graph Lifelong \\ Learning\end{tabular}} \\ \hline
Graph Representation Learning & \multicolumn{1}{c|}{Yes} & \multicolumn{1}{c|}{Yes} & \multicolumn{1}{c|}{Yes} & Yes \\ \hline
Online Learning & \multicolumn{1}{c|}{No} & \multicolumn{1}{c|}{Yes} & \multicolumn{1}{c|}{Yes} & Yes \\ \hline
Transfer Knowledge & \multicolumn{1}{c|}{No} & \multicolumn{1}{c|}{No} & \multicolumn{1}{c|}{Yes} & Yes \\ \hline
Incremental Learning & \multicolumn{1}{c|}{No} & \multicolumn{1}{c|}{No} & \multicolumn{1}{c|}{No} & Yes \\ \hline
Knowledge Retention & \multicolumn{1}{c|}{No} & \multicolumn{1}{c|}{No} & \multicolumn{1}{c|}{No} & Yes \\ \hline
Citation & \multicolumn{1}{c|}{\cite{contribute_wu_GNNSurvey_2019,contribute_feng_graphSurvey_2021,contribute_zhang_graphSurvey_2020}} & \multicolumn{1}{c|}{\cite{overviewSpatiojainStructuralrnnDeepLearning2016,method_stgcn_yu2017,overviewSpatiozhangGaanGatedAttention2018, spatio_attention_guo2019}} & \multicolumn{1}{c|}{\cite{relatedSurveykazemiRepresentationLearningDynamic2020,relatedSurveybarrosSurveyEmbeddingDynamic2021, overviewDynamicxueDynamicNetworkEmbedding2022,dynamic_dyngem_goyalDyngemDeepEmbedding2018,dynamic_dyngraph2vec_GOYAL,dynamic_triad_zhouDynamic2018}} & {\cite{method_FGN_wang_2020, method_ergnn_zhou_2020, benchmark_carta2021catastrophic}} \\ \hline
\end{tabular}
}
\end{table*}
\subsection{Relevant ML concepts}\label{subsection:ML}
This section discusses some learning paradigms in typical machine learning models, such as transfer learning, multitask learning, and online learning \cite{chen_liu_2018_LifelongMachineLearning} and highlights how they differ from lifelong learning. Transfer learning seeks to take the knowledge gained from learning a task in a particular source domain and transfer it over to a different, but related, target domain \cite{zhuang2020comprehensiveTransfer}. The transfer learning paradigm has no specific mechanism designed to retain past knowledge and use it to learn new knowledge, which is the main goal of lifelong learning. Multitask learning aims to share knowledge across various tasks to help the learning process \cite{zhang2021surveyMultitask}. Instead of optimizing a single task, multitask learning aims to maximize the performance of the tasks simultaneously. Data is assumed to be static and the tasks are predefined, and there is no mechanism to accommodate new knowledge that comes sequentially. Lastly, online learning deals with future data in sequential order but has a different objective than lifelong learning. Online learning aims to learn similar and predefined tasks sequentially to achieve efficiency \cite{hoi2021online}. On the other hand, lifelong learning aims to learn new knowledge while retraining previous experience and using that to help with future tasks. 

Another concept of some relevance is out-of-distribution (OOD) generalization techniques~\cite{yang2021generalized_OOD, li2022out_graphood}. OOD generalization aims to create a model that can perform well on test distributions that differ from training distributions. The target of that concept is to produce a robust predictor based on invariant features that still have a similar domain~\cite{lesort2021understanding_continual_OOD}. This is different from the objectives of lifelong learning, however, which aims to accommodate new data distributions with different contexts from existing ones, whilst trying to avoid shifting performance or forgetting previous knowledge learned from past data distributions.

\subsection{Relevant Graph Learning Characteristics}
\label{rlp}
In order to utilize graph data in learning algorithms, the first strategy is to capture the graph representation and convert it into a meaningful feature representation. This is known as graph representation learning. It is common in graph learning problems to assume the graph is static and models a fixed number of tasks in terms of classes and data instances before the training process begins. There are, however, specific fields within graph learning that seek to address dynamic data. Spatio-temporal graph networks utilize dynamic graph learning over time to account for dynamic temporal attributes where the structure of the graph remains static. Yu et al. \cite{method_stgcn_yu2017} proposed a novel framework called spatio-temporal graph convolutional networks (STGCN) to develop time-series prediction in the traffic domain. There are two essential components in this type of spatio-temporal graph. The first is spatial forms that maintain the static structure of the graph (e.g., in \cite{method_stgcn_yu2017} the traffic network). The second component is temporal features that describe different values of relations between nodes that change over time (e.g., sensor readings/values at the node). The objective of this method is to perform convolutional operations on a graph to utilize spatial information to forecast the graph edges' temporal features. Guo et al. \cite{spatio_attention_guo2019} developed a novel attention-based spatial-temporal graph convolutional network (ASTGCN) to address the same objective of forecasting traffic situations. The main contribution of the ASTGCN model is to employ spatial attention to consider the influence of other nodes on traffic conditions. This method uses the same approach of spatial-temporal convolutions to obtain valuable information on the spatial form and temporal features graph. Although those methods incorporate dynamic temporal attributes in the graph, they have different characteristics from the concept of lifelong learning. The embedding processes on those methods are carried out sequentially at different timeframes, and there is no knowledge transfer mechanism when a new node representation appears in the graph. Moreover, spatio-temporal neural network techniques have predefined and fixed tasks that do not consider accommodating new knowledge, so the concept of addressing the problem of catastrophic forgetting is not discussed in this learning concept.

Another graph learning approach that shares some similarities with graph lifelong learning is dynamic graph learning which can be called temporal graph embedding \cite{yang2021discrete}. Dynamic graph learning can be divided into two graph types, a discrete-time evolving graph which can be represented as a collection of dynamic graph snapshots at different time steps, and a continuous-time evolving graph, which carries more information and complexity and is defined as a graph stream instead of snapshots \cite{skarding2021foundations}. Some works address discrete-time evolving graphs, such as a work by Goyal et al. \cite{dynamic_dyngem_goyalDyngemDeepEmbedding2018} called DynGEM, based on deep autoencoders that aim to perform dynamic embedding in graphs in discrete time with snapshots. DynGEM uses the learned embedding from a previous time step to initialize the current time step embedding so as to transfer knowledge and ensure the embedding remains close to that of the previous timestep. Zhou et al. \cite{dynamic_triad_zhouDynamic2018} proposed a novel dynamic network embedding approach called DynamicTriad, which aims to learn a representation of dynamic graphs by imposing a triad (group of three vertices) to learn the representation while also preserving temporal information. These dynamic graph learning methods share graph lifelong learning's goal of transferring past knowledge to produce a good representation of consolidated knowledge in the next time step to help the model learn efficiently. 

For continuous-time-evolving graphs, Ma et al. \cite{ma2020streaming} proposed a method called streaming graph neural networks (SGNN) that aims to do strict representation learning on continuous dynamic node representation. The architecture of SGNN consists of two components, an update component to modify the interaction information of new relationships and a propagation component to propagate the update to the involved or relevant neighborhood based on a time-aware long-short-term memory (LSTM). These dynamic graph learning approaches for both discrete and continuous networks aim to perform representation learning based on every change in graph snapshots or graph streams. However, there are some noticeable differences between dynamic graph learning and graph lifelong learning. Firstly, dynamic graph learning only focuses on the embedding process in the graph without considering the addition of downstream tasks, such as the addition of new classes or data instances to tasks, which are also very important and relevant to the concept of incremental learning. Second, those models do not consider performing backward transfers or refining prior embedding of graph representations when learning new knowledge. In particular, there is no precise mechanism to address catastrophic forgetting problems, which is the main challenge in lifelong learning.

The different characteristics of the graph learning methods described above are summarized in Table \ref{conceptCompare}. The following are properties that are related to graph lifelong learning:

\begin{itemize}
  \item \textbf{Graph representation learning}. The model aims to learn the representation of irregular domains, \emph{e.g.}, graphs, and translate it into feature vectors to solve downstream graph-based tasks.
  \item \textbf{Online learning}. The model is able to learn in the continuous data stream.
  \item \textbf{Knowledge transfer}. The model can help future learning by transferring previous knowledge as initial knowledge for the new tasks.
  \item \textbf{Incremental learning}. The model should be able to learn incrementally for new upcoming tasks in terms of new instances or new classes in the future time points without training from scratch and relying on previous data as much as possible.
  \item \textbf{Knowledge retention}. The model should be sturdy from catastrophic forgetting problems and able to refine previous experiences when learning new knowledge.
\end{itemize}

\subsection{Graph Lifelong Learning Scenarios}
\label{secenarios}
Maltoni and Lomonaco~\cite{motive_categorization_maltoni_2019} introduced three scenarios that are relevant to class-incremental learning: learning new instances, new classes, and new instances and classes. Here we explain the content update types in the graph lifelong learning setting:

\begin{itemize}
  \item \textbf{Learning new instances (NI)}. This refers to the case where new training patterns appear in future time points based on new conditions, but without the addition of new classes. In graph learning problems, this can be indicated by the emergence of new nodes and their attributes that are completely different from the existing nodes. Thus, it can create new relationships in graph data. It is necessary to accommodate the new information of newly appeared nodes and relations to improve the scalability of models.
  
  \item \textbf{Learning new classes (NC)}. This is when new training patterns that correspond to previously unseen classes become available. Once new classes appear, the model should be able to accommodate the new information to solve the tasks based on the new classes. Moreover, previous experiences should be maintained.
  
  \item \textbf{Learning new instances and classes (NIC)}. New training patterns and new labels become available in future observation. Newly emerged data instances can belong to both previous classes or new classes that can improve the prior experience and enrich the model capability based on new knowledge.
  
\end{itemize}
As mentioned by Maltoni and Lomonaco~\cite{motive_categorization_maltoni_2019}, NI and NIC are often more appropriate than NC in real-world applications.

\subsection{Notation of Domain}
\label{sec:notationDomain}
A graph is composed of a set of vertices (or nodes) $V$ containing attributes with a total number of vertices denoted as $|V|=n$ , and a set of edges $E$. A vertex or a node can be represented as $v_i \in V$, and an edge to define a relation can be represented as $e_{ij} = (v_i,v_j) \in E$ which determines the relationship between vertices $v_i$ to $v_j$. Graph data can be represented as $G = (A, X)$, where $A \in \{1,0\}^{n \times n}$ is the adjacency matrix and $X \in \mathbf{R}^{n \times d}$ represents features in nodes where $n$ is the number of nodes, and $d$ is the dimension of a feature vector. In graph lifelong learning setting, the change of graph data based on time $t$ can be represented as $G = (G_1, G_2, ..., G_t)$ where $G_t = G_{t-1} + \Delta G_t$. Moreover, the graph lifelong learning problem will have a collection of tasks $T=(T_1,T_2,...,T_t)$.

\textit{\textbf{Definition 1.} The objective of graph lifelong learning is to learn a task $T_t$ with graph data  $G_t$ that comes sequentially by maximizing the result of the model $f$ (parameterized by $\theta$) while maintaining the performance and avoiding the catastrophic forgetting problem for tasks $T_1,T_2,...,T_{t-1}$.}

\begin{table*}
\centering
\caption{Graph Lifelong Learning Method Comparison}
\resizebox{\textwidth}{!}{
\begin{tabular}{|l|l|l|l|l|} \hline
\multicolumn{1}{|c|}{\textbf{Methods}} & \multicolumn{3}{c|}{\textbf{Approaches}} & \multicolumn{1}{|c|}{\textbf{Code Links}} \\ \cline{2-4}
\multicolumn{1}{|c|}{} & \multicolumn{1}{c|}{\textbf{Architectural}} & \multicolumn{1}{c|}{\textbf{Rehearsal}} & \multicolumn{1}{c|}{\textbf{Regularization}} &  \\ \hline
\begin{tabular}[c]{@{}l@{}}Feature Graph Networks\\(FGN) \cite{method_FGN_wang_2020}\end{tabular} & Yes & No & No & https://github.com/wang-chen/LGL \\ \hline
\begin{tabular}[c]{@{}l@{}}Hierarchical Prototype Networks~\\(HPNs) \cite{method_HPNs_zhang_2021}\end{tabular} & Yes & No & No & - \\ \hline
\begin{tabular}[c]{@{}l@{}}Experience Replay GNN\\Framework (ER-GNN) \cite{method_ergnn_zhou_2020}\end{tabular} & No & Yes & No & - \\ \hline
\begin{tabular}[c]{@{}l@{}}Lifelong Open-world \\Node Classification (LONC) \cite{method_openWorld_2021}\end{tabular} & No & Yes & No & https://github.com/lgalke/lifelong-learning \\ \hline
\begin{tabular}[c]{@{}l@{}}Disentangle-based Continual graph \\Representation learning (DiCGRL) \cite{method_DiCGRL_kou_2020}\end{tabular} & No & No & Yes & https://github.com/KXY-PUBLIC/DiCGRL \\ \hline
\begin{tabular}[c]{@{}l@{}}Graph Pseudo Incremental \\Learning (GPIL) \cite{method_GFCIL_zheng_2022} \end{tabular} & No & No & Yes & https://github.com/zhen-tan-dmml/gfcil \\ \hline
\begin{tabular}[c]{@{}l@{}}Topology-aware Weight Preserving\\(TWP) \cite{method_TWP_liu_2020} \end{tabular} & No & No & Yes & https://github.com/hhliu79/TWP \\ \hline
\begin{tabular}[c]{@{}l@{}}Translation-based Knowledge Graph \\Embedding (TKGE) \cite{method_translationBased_song_2018} \end{tabular} & No & No & Yes & - \\ \hline
\begin{tabular}[c]{@{}l@{}}ContinualGNN \cite{method_continualGNN_wang_2020}\\\\\end{tabular} & No & Yes & Yes & https://github.com/Junshan-Wang/ContinualGNN \\ \hline
\begin{tabular}[c]{@{}l@{}}Lifelong Dynamic Attributed \\Network Embedding (LDANE) \cite{method_LDANE_wei_2019}\end{tabular} & Yes & Yes & Yes & - \\ \hline
\begin{tabular}[c]{@{}l@{}}TrafficStream \cite{method_trafficStream_chen_2021}\\\\\end{tabular} & No & Yes & Yes & https://github.com/AprLie/TrafficStream \\ \hline
\end{tabular}
}
\end{table*}
\subsection{Taxonomy}
\label{sec:taxo}
\begin{figure}[h]
	\begin{center}
		\includegraphics[scale=0.745]{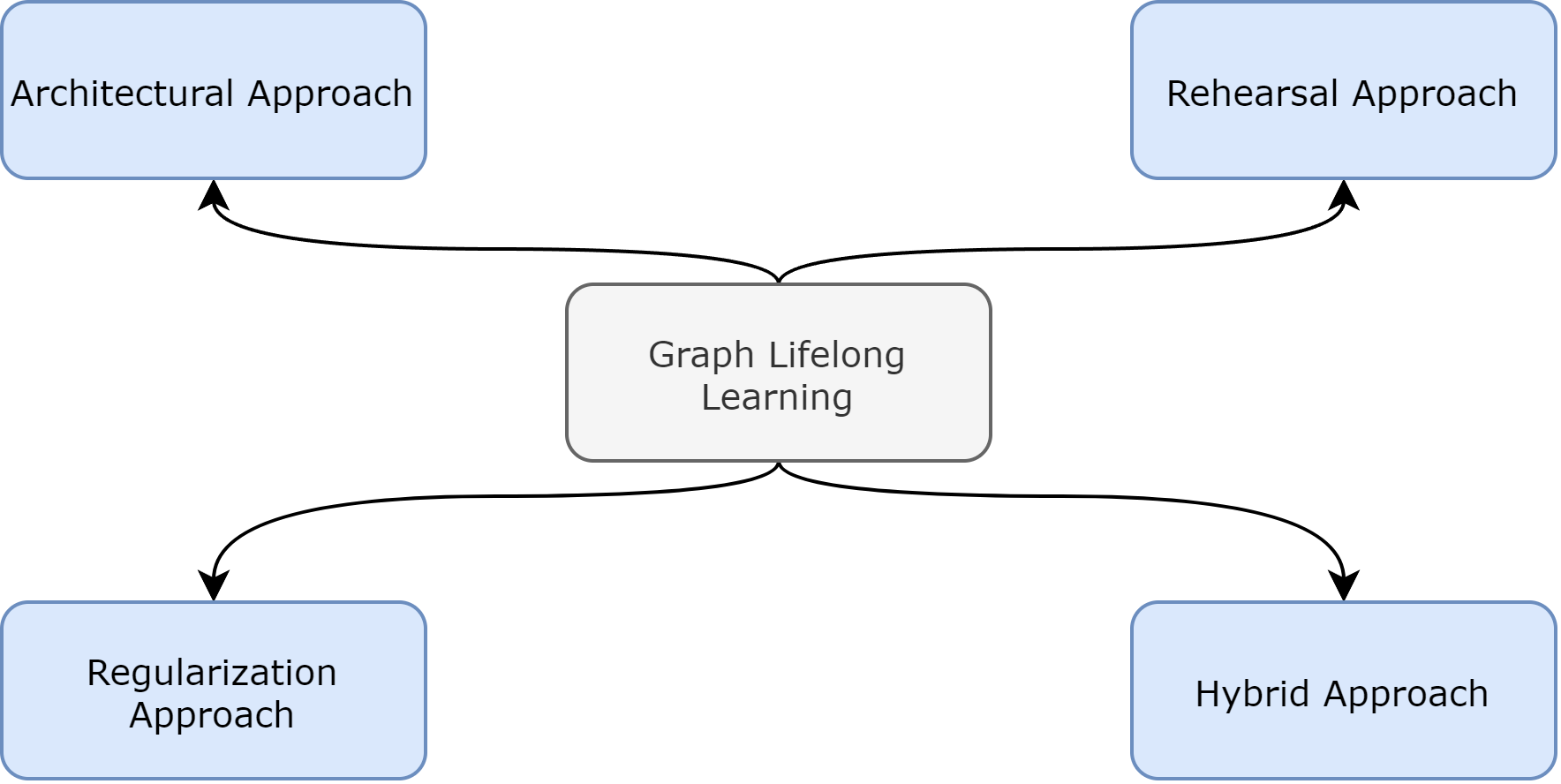}
	\end{center}
	\caption{Graph Lifelong Learning Categorization.}
	\label{modelframework}
\end{figure}

This section presents a categorization of graph lifelong learning research. The categorization is based on the one presented for general lifelong learning, \cite{motive_categorization_maltoni_2019, motive_categorization_biesialska_2020_survey}, as explained in Subsection \ref{sec:life}, which we have found is also relevant to use for graph lifelong learning. The unique considerations in the case of graph lifelong learning are how these methods consider graph structure data in the incremental learning process. The hybrid approach is also added to categorize the current methods that combine more than one approach. Thus, the existing graph lifelong learning works covered in this survey paper are divided into four categories: architectural, regularization, rehearsal, and hybrid, as shown in Fig. \ref{modelframework}.

\section{Architectural Approach}
\label{arch}
In the general lifelong learning setting, this approach focuses on modifying the specific architecture of networks, activation functions, or layers of algorithms to address a new task and prevent the forgetting of previous tasks \cite{motive_definitionLifelong_ruvolo_ELLA, method_progNN_rusu_2016_, method_cwr_lomonaco_2017}. In the graph lifelong learning scenario, this approach relies on changing the graph structure, expanding more units, and performing compression techniques. Some examples are FGN \cite{method_FGN_wang_2020}, which convert graph data architecture into regular learning problems, and HPNs~\cite{method_HPNs_zhang_2021}, which extract different level abstractions of prototypes to accommodate new knowledge. 

\subsection{Feature Graph Networks}
In a graph learning process, the full adjacency matrix or the entire graph topology information is needed to propagate node features across all layers. Sampling techniques have been proposed to scale GCNs to large graphs \cite{bench_reddit_hamilton2017inductive, chen2018fastgcn}, however GCNs still require a costly pre-processing step that includes the entire graph dataset, which is not suitable for incremental learning in lifelong learning. In particular, the topology of graphs provides unique challenges for lifelong learning. This has motivated the proposal of a new graph topology that is suitable for general lifelong learning approaches.

\begin{figure}[h]
	\begin{center}
		\includegraphics[scale=0.3]{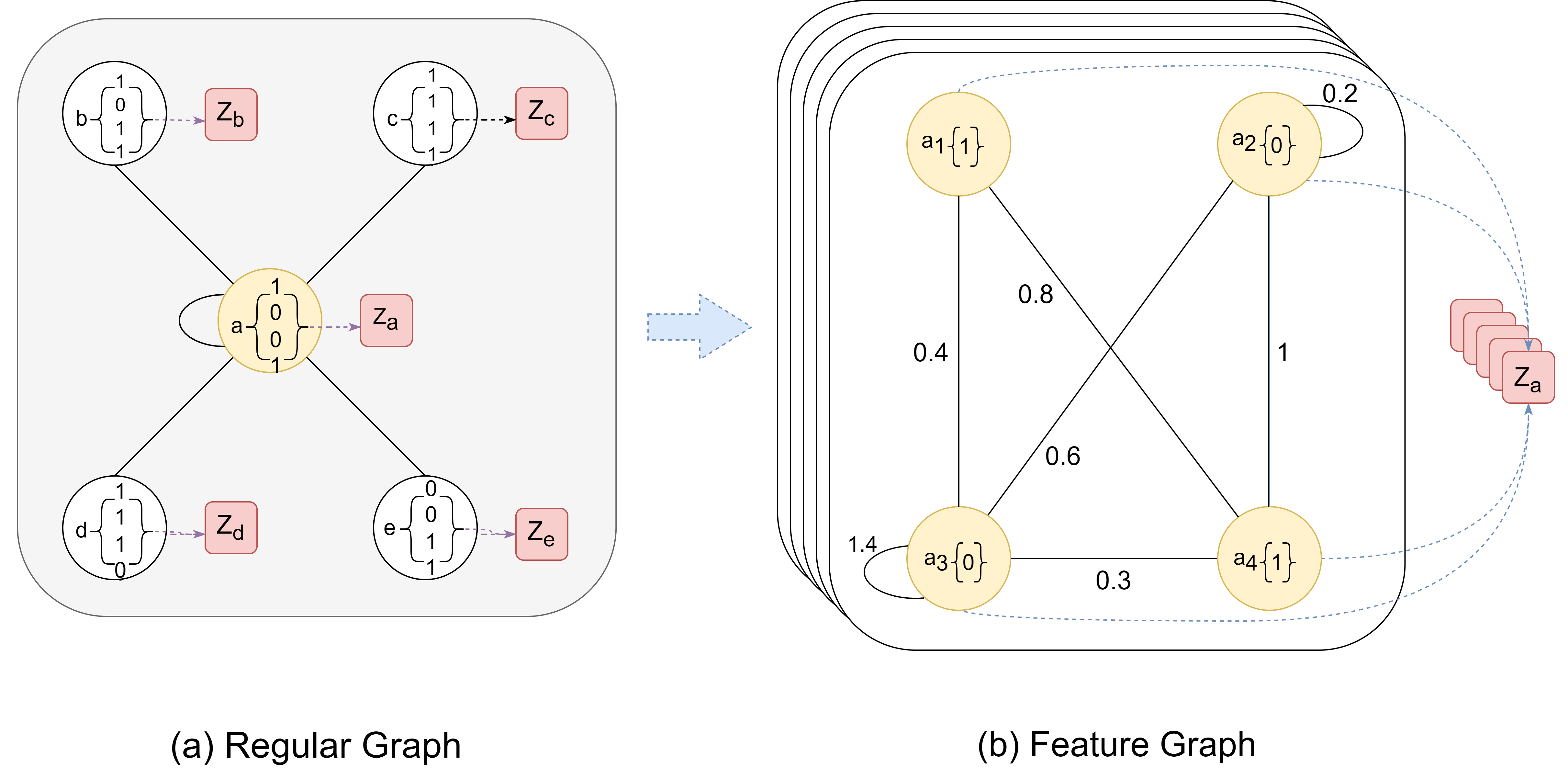}
	\end{center}
	\caption{Illustration of FGN. It changes the structure of graphs. In the regular graph representation shown in (a), the node $a$ has neighbors $N(a)=\{a,b,c,d,e\}$ and feature vector $x_a = [1,0,0,1]$. FGN converts the regular graph so that the features of node $a$ will be represented as nodes $\{a_1,a_2,a_3,a_4\}$ in the Feature Graph shown in (b), connected to each other via cross-correlation establishment.   \cite{method_FGN_wang_2020}.}
	\label{fgn}
\end{figure}

Wang et al. \cite{method_FGN_wang_2020} developed the FGN approach to convert graph architectures into a representation that can be ingested by regular learning architectures. Figure \ref{fgn} shows how FGN converts regular graph representations into feature graphs. In particular, FGN represents nodes from the original graph as independent graphs, and the features from the original graph as nodes. This technique enables lifelong learning approaches based on CNNs architectures to be applied to solve GNN tasks.

The FGN architecture modification allows graph problems to be re-framed, by transforming a node classification task into a feature graph classification task. This then enables the feature graphs to be trained in a regular mini-batch manner similar to CNNs \cite{methods_FGNminibatch_kri_2017}. By converting graphs into a representation that suits common CNNs tasks, FGN enables the input of the graph's node to be processed independently whilst new knowledge is being learned, without requiring the entire graph to be pre-processed. This also changes some graph relationships: fixed feature dimensions in the regular graph translate to a fixed number of nodes in the feature graph, and the addition of nodes to the adjacency matrix of the regular graph translates to the addition of adjacency matrices of feature graphs. The cross-correlation of connected feature vectors is accumulated to develop feature interactions that are useful for constructing feature adjacency. Moreover, in order to develop a feature graph that is a new topology given by the feature adjacency matrix, specific types of layers can be used to develop the feature graph, such as the feature broadcast layer, feature transform layer, and feature attention layer. 

One of the methods to calculate the performance of this model uses data-incremental and class-incremental scenarios. The experiments use several datasets such as Cora~\cite{bench_coraCiteseerPubmed_sen2008}, Citeseer~\cite{bench_coraCiteseerPubmed_sen2008}, Pubmed~\cite{bench_coraCiteseerPubmed_sen2008}, and Ogbn-arXiv~\cite{bench_arxiv_mikolov2013distributed,bench_arxiv_wang2020microsoft}. FGN is compared with other graph models such as GCN~\cite{motive_GCN_kipf_2016}, graph attention networks (GAT)~\cite{dicgrl_ne_GAN_velivckovic2017}, graphSage~\cite{bench_reddit_hamilton2017inductive}, and approximation personalized propagation of neural predictions (APPNP)~\cite{klicpera2018predict_APPNP} that are implemented for lifelong learning settings without knowledge consolidations.  On-data incremental scenarios, all new samples are randomly inserted. The result shows to get 0.872 $\pm$ 0.009 of overall performance on the Pubmed dataset. In the class-incremental scenario, all samples from each class are inserted into the model before moving to another class, and it gets 0.857 $\pm$ 0.003 of performance on the Pubmed dataset. Moreover, it gets a forgetting rate value of 3.49, which is the lowest forgetting rate compared to the other models. It shows effectiveness by achieving superior performance in both data-incremental and class-incremental tasks.

\subsection{Hierarchical Prototype Networks}
Hierarchical Prototype Networks by Zhang et al. \cite{method_HPNs_zhang_2021} extract different levels of knowledge abstraction in the form of prototypes, which are used to represent the evolving graph over time. This method aims to accommodate incoming experience whilst retaining prior knowledge. It firstly leverages a set of atomic feature extractors (AFEs) to encode attribute information and topological structures of target nodes. There are two kinds of atomic embeddings extracted in this process. The first one is atomic node embedding $AFE_{node}$ to generate an embedding representation of node $v$'s feature information,  $\mathbb{E}^{node}_A(v)$. The second one is atomic structure embedding $AFE_{struct}$. The atomic structure embedding of node $v$, denoted $\mathbb{E}^{struct}_A(v)$, encodes the relations between nodes within multiple hops of node $v$. Then, HPNs perform several functions to adaptively select, compose, and store the embedding representation through three architecture layers of prototypes: atomic-level, node-level, and class-level. First, the atomic-level prototypes (A-prototypes) $\mathbb{P}_A$ are created by atomic embedding sets, which describe low-level features of a node that is similar to the output of the first layer in deep neural networks. Then node-level (N-prototypes) $\mathbb{P}_N$ and class-level prototypes (C-prototypes) $\mathbb{P}_C$ are developed to capture high-level information of nodes. N-prototypes $\mathbb{P}_N$ are developed by capturing the relation of each node in A-prototypes $\mathbb{P}_A$. Then, C-prototypes $\mathbb{P}_C$ are generated from N-prototypes $\mathbb{P}_N$ to explain the common feature of a node group. All those conversions use the feature map of convolutional neural networks at different layers.

HPNs choose only relevant atomic embeddings to refine existing prototypes or create new prototypes to maintain old knowledge and accommodate new knowledge. HPNs use maximum cosine similarity to calculate the similarity between incoming atomic embedding and atomic embedding on existing prototypes to select the relevant knowledge. The result of maximum cosine similarity is then ranked and taken according to the number specified in the hyperparameters. The HPNs then determine which new embedding has the same or different characteristics as the elements in the A-prototypes $\mathbb{P}_A$ using cosine similarity. On one hand, when new embeddings are considered close to existing prototypes, the old knowledge will be refined by minimizing the loss between the two. On the other hand, atomic embeddings that are not close to the atomic embedding of existing prototypes will be considered as new tasks, so a new prototype will be degenerated by combining previous and new prototypes. In the scenario that new embeddings $\mathbb{E}_{new}(v)$ have similar representation to the previous representation, that can cause redundant knowledge. In order to avoid this, the model filters $\mathbb{E}_{new}(v)$ to get $\mathbb{E}^\prime_{new}(v)$ which only contain distinct representations (as measured using the threshold $t_A$):

\begin{equation}
    \forall e_i,e_j \in \mathbb{E}^\prime_{new}(v), \frac{e^T_ie_j}{\|e_i\|_2\|e_j\|_2}<t_A.   
\end{equation}

Using those strategies, the embeddings in $\mathbb{E}^\prime_{new}(v)$ will only contain new features. Then the A-prototypes $\mathbb{P}_A$ are updated to accommodate new embeddings.

\begin{equation}
    \mathbb{P}_A = \mathbb{P}_A \cup \mathbb{E}^\prime_{new}(v).
\end{equation}

After that process, the selected A-prototypes $\mathbb{P}_A$ can be further matched to $\mathbb{P}_N$ and C-prototypes $\mathbb{P}_C$ to get hierarchical representation, which is used to feed the classifier to perform node classification. 

The architectural approach to HPNs that activates hierarchical representative embedding and iterative prototypes to accommodate new knowledge achieves state-of-the-art performance based on accuracy mean and forgetting mean to enable lifelong learning in graph data. It is compared with general lifelong learning, such as EWC \cite{method_ewc_kirkpatric_2016}, LWF \cite{method_lwf_li_2016}, and gradient episodic memory (GEM) \cite{method_GEM_lopez2017gradient}. Those methods are combined directly with three different base methods of graph representation learning without any knowledge consolidation, such as GCNs \cite{motive_GCN_kipf_2016}, GAT~\cite{dicgrl_ne_GAN_velivckovic2017}, and graph isomorphism network (GIN) \cite{xu2018powerful}. The scenarios are to learn new tasks based on adding new data and classes in some datasets such as Cora~\cite{bench_coraCiteseerPubmed_sen2008}, Citeseer~\cite{bench_coraCiteseerPubmed_sen2008}, Actor~\cite{pei2020geom_datasetActor}, etc. The result shows that HPNs get the best average accuracy of 93.7 $\pm$ 1.5 in the Cora dataset. The forgetting rate also gets the least rate compared with other methods, with -0.9 $\pm$ 0.9 in the Actor dataset. HPNs not only perform better in terms of performance but also have good memory efficiency.

\section{Rehearsal Approach}
\label{rehear}
This approach regulates retraining processes for previous tasks to strengthen the relationship between memory and performance on previously learned tasks \cite{method_icarl_rebuffi_2017, method_exstream_hayes_2019}. This graph learning approach enables the selection of appropriate samples of graph representation, such as nodes and edges, for retraining purposes. The number of samples is carefully considered to minimize the computational complexity. Recently, graph lifelong learning methods using rehearsal approaches have been proposed, such as ER-GNN~\cite{method_ergnn_zhou_2020}, and lifelong Open-world Node Classification \cite{method_openWorld_2021}. Other graph lifelong learning approaches, such as ContinualGNN \cite{method_continualGNN_wang_2020} and TrafficStream \cite{method_trafficStream_chen_2021}, that combine aspects of the rehearsal approach with other different approaches are discussed in Section \ref{hybrid}.

\subsection{Experience Replay GNN Framework}
Zhou et al. \cite{method_ergnn_zhou_2020} proposed the concept of lifelong learning on graph data using graph-based experience replay called ER-GNN. The main objective is to mitigate the catastrophic forgetting problem in graph data while learning continuously over time. The model preserves past knowledge in an experienced buffer obtained from previous learning of tasks to be replayed while learning a new task. This model implements an experience node strategy to help to select the experience nodes to be stored in the experience buffer, such as the mean of feature (MF) \cite{method_icarl_rebuffi_2017} and coverage maximization (CM) \cite{method_ergnnCoverage_tim_2016}. Apart from those experience node strategies, ER-GNN proposed a new method called influence maximization (IM).

The model's experience node replay is influenced by Complementary Learning System (CLS) that supports the biological learning process in humans. CLS is a great example of a complementary contribution to the virtual experience mechanism of the hippocampus and neocortex in the human brain system. It helps consolidate knowledge by replaying memories in the hippocampus before being transferred into long-term memory using the neocortical system \cite{method_ergnncls_kumaran_2016}. Referring to the ER-GNN problem definition, suppose there are a collection of tasks $T = \{T_1,T_2,...,T_i,...T_M\}$ which come sequentially and each $T_i \in T (|T| = M)$. Every task $T_i$ has training node-set $D^{tr}_i$ and testing node-set $D^{te}_i$. When the model starts to learn task $T_i$, it uses resources from the training set $D^{tr}_i$ and maximizes the performance using the testing set $D^{te}_i$. It also selects the experience of examples $B$ from the stored experience buffer. Both the training data $D^{tr}_i$ and experience nodes $B$ help to learn a model $f_\theta$ parameterized by $\theta$. ER-GNN uses a cross-entropy loss function for node classification and implements parameter updates to get optimal parameters by minimizing the loss using optimization methods such as Adam optimizer \cite{method_ergnnAdam_kingma_2015}. In the next step, after updating the parameters, the model performs node selection in the current training data set $D^{tr}_i$ to be added as an example into the experience buffer. MF, CM, and the novel method, IM are used as a scheme based on the crucial part of the experience selection strategy that stores experience in the buffer to increase the performance of continuous graph learning. 

To measure the performance of ER-GNN, it uses a case study to learn new tasks based on additional new nodes from new classes to perform node classification. Several datasets, such as Cora~\cite{bench_coraCiteseerPubmed_sen2008}, Citeseer~\cite{bench_coraCiteseerPubmed_sen2008}, and Reddit~\cite{bench_reddit_hamilton2017inductive}, are used in this evaluation. Some graph learning models, such as Deepwalk~\cite{perozzi2014deepwalk}, Node2Vec~\cite{grover2016node2vec}, GCN~\cite{motive_GCN_kipf_2016}, graphSage~\cite{bench_reddit_hamilton2017inductive}, etc., are used in comparison with a finetuning strategy which means the model is trained continually without employing lifelong learning techniques. The result shows ER-GNN achieves the best mean performance on the Cora dataset with mean accuracy of 95.66\%. It also has the lowest mean forgetting, 17.08\%, compared to graph learning with finetuning strategy. From that experiment, it shows the performance improvement in reducing forgetting mean compared to state-of-the-art GNNs that are forced to do learning a sequence of node classification tasks.

\subsection{Lifelong Open-world Node Classification}
Galke et al. \cite{method_openWorld_2021} introduced lifelong open-world classification to implement incremental training on graph data. This model could rely on previous knowledge to help the learning process in a sequence of tasks. The learned knowledge is stored as historic data explicitly or within model parameters implicitly. The model then analyses the influence of stored knowledge that is helpful in refining the network. The model's objective is to address the challenge in graph learning that nodes in graph data can not be processed independently because the relation of nodes needs to be processed through an embedding mechanism to share the information across the nodes. The second objective is to enable the concept of open-world classification \cite{chen_liu_2018_LifelongMachineLearning} to implement incremental training in graph data and accommodate new tasks that are completely different from the previous ones. The model introduces a new measurement to regulate the evolution in graph data with different characteristics to address that objective. The method has the capability to integrate an isotropic and anisotropic type \cite{method_openworldIsotro_dwivedi_2020} of GNN.

The problem definition of this model is as follows. Assuming graph data has a finite number of tasks $T_1,...,T_t$ and a model $f$ with parameters $\theta$. To do incremental training in this method, while the structure of the graph including its nodes and edges is changing over time, during preparation for task $T_{t+1}$, a model should have been trained based on the label on task $T_t$ to get the $ \theta^t)$ parameters. When $l$ new classes come, these methods add the corresponding number of parameters to the output layer of $f^t)$. Therefore it will have a new parameterized layer:

\begin{equation}
\mid\theta^{t}_{output\,weight}\mid = \mid\theta^{t-1}_{output\,weight}\mid + \l, \text{and} 
\end{equation}

\begin{equation}
\mid\theta^{t}_{output\,bias} \mid = \mid\theta^{t-1}_{output\,bias}\mid + \l. 
\end{equation}

In the next step, to enable the rehearsal approach, this method gives the option to choose whether to retrain the model from scratch (cold restart) or use the final model parameters from the previous tasks (warm restart). The model involves an artificial history size that determines the amount of past data to control against full-history retraining.

Since the pre-compiled dataset for graph lifelong learning is rare, this model performs an experiment on the scientific publication dataset PharmaBio and DBLP (easy and hard)~\cite{tang2008arnetminer_dblbDataset}. The scenario is to learn new tasks based on adding new classes. Two evaluation measures used average accuracy and forward transfer to measure the effect of reusing previous parameters. It also uses several graph neural networks technique to perform representation learning in the model, such as GAT~\cite{dicgrl_ne_GAN_velivckovic2017}, GCN~\cite{motive_GCN_kipf_2016}, and graphSage~\cite{bench_coraCiteseerPubmed_sen2008}. The main result shows that incremental learning using limited history sizes is almost similar to the default setting by using the full history size of the graph. GNN technique in this model achieves at least 95\% accuracy that can be retrained with half dataset coverage compared to using all past data for incremental learning.

\section{Regularization Approach}
\label{regu}
This approach implements a single model and has a fixed capacity by leveraging the loss function using the loss term to help consolidate knowledge in the learning process for new tasks and retain previous knowledge \cite{method_ewc_kirkpatric_2016, method_lwf_li_2016}. Prior knowledge of graph structures and tasks will be maintained to achieve stable performance while learning novel knowledge. In graph-based learning, some current works that implement regularization approaches are DiCGRL~\cite{method_DiCGRL_kou_2020}, GPIL~\cite{method_GFCIL_zheng_2022}, TWP~\cite{method_TWP_liu_2020}, and Translation-based Knowledge Graph Embedding \cite{method_translationBased_song_2018}. Other methods that partially implement a regularization approach in graph data that are explained in more detail in Section \ref{hybrid} are ContinualGNN \cite{method_continualGNN_wang_2020}, LDANE \cite{method_LDANE_wei_2019} and TrafficStream \cite{method_trafficStream_chen_2021}.

\subsection{Disentangle-based Continual Graph Representation learning}
Kou et al. \cite{method_DiCGRL_kou_2020} proposed Disentangle-based Continual Graph Representation Learning (DiCGRL), which focuses on applying lifelong learning in graph embedding. Current graph embedding models are not practical in real-world applications because they tend to ignore streaming and incoming data nature in graphs. DiCGRL takes two approaches to continuously learn a new graph embedding whilst preventing the forgetting of old knowledge. First, the disentangle module aims to convert the graph's relational triplets ($u$,$r$,$v$), which explains the connection between node $u$ and $v$ based on relation $r$ into several components in their semantic aspects. The model uses an attention mechanism to learn the attention weight of components $u$ and $v$ based on the relation $r$. Then, top-n-related components are processed through a graph embedding mechanism that incorporates the features of nodes $u$ and $v$. There are two types of graph embedding used to do this: knowledge graph embedding (KGEs) \cite{dicgrl_kgs_rotate_sun2019G,dicgrl_kgs_transe_bordes2013} and network embedding (NEs)\cite{dicgrl_ne_GAN_velivckovic2017, dicgrl_ne_DGCN_henaff2015}.

The second step is updating the module that aims to update disentangled graph embedding when new relation triplets appear. This model uses a regularization approach to prevent the forgetting problem. In this process, a step called neighbor activation identifies which triplet relation should be updated by running a selection mechanism inspired by the human ability to learn procedural knowledge. Top-n components with common characteristics are selected to update the embedding representation. Then, when the new data arrive, the incoming data will be trained with those activated neighbor components using a constraint in loss function to accommodate new knowledge and maintain the performance of prior tasks.
DiCGRL employs constraint loss terms $\mathcal{L}_{norm}$ to push the sum of attention weights of top-n selected components reaching 1, \emph{i.e.},

\begin{equation}
\mathcal{L}_{norm} = \sum_{(u,r,v) \epsilon \mathcal{T}_i} (1 - \sum_k^n \alpha^k_r), \end{equation}
where $n$ indicates the number of components. So that the overall loss function to implement the regularization process in this method is as follows:

\begin{equation}
\mathcal{L} = \mathcal{L}_{old} + \mathcal{L}_{new} +\beta . \mathcal{L}_{norm},
\end{equation}
where $\beta$ is a hyperparameter.

One of the techniques to measure the performance of DiCGRL by using link prediction tasks and node classification tasks. For link prediction tasks, it uses two datasets, FB15K-237~\cite{toutanova2015observed_FBdataset} and WN18RR ~\cite{dettmers2018convolutional_WNdataset}, that are set to have instances that come sequentially. Moreover, two types of knowledge graph embedding, such as TransE~\cite{dicgrl_kgs_transe_bordes2013} and ConvKB~\cite{nguyen2017novel_convkb}, are used to perform graph representation learning. For node classification tasks, It uses several datasets such as Cora, CiteSee, and Pubmed~\cite{bench_coraCiteseerPubmed_sen2008} with the scenario of adding instances sequentially. There are some strategies used to compare the performance of the method. First, as a lower bound, finetuning is implemented to train the data continuously without employing lifelong learning. Second, as the upper bound, all nodes are retrained from scratch when new information appears. Moreover, there are some general lifelong learning that does not consider graph representation learning, such as EWC~\cite{method_ewc_kirkpatric_2016}, EMR~\cite{motive_lifelongDefinition_parisi_2019}, and GEM~\cite{method_GEM_lopez2017gradient}. For link prediction, DiCGRL achieves the maximum average performance of H@ 10 (average top-10 ranked entities) on the FB15K-237 dataset with 47.7\%. The highest average accuracy on the Pubmed dataset for node classification is 85\%. DiCGRL achieves an improvement compared to the general lifelong learning techniques and tends to have a close result with the upper bound baselines.

\subsection{Graph Few-shot Class-incremental Learning}
The ability to learn new classes incrementally is essential in real-world scenarios. Zhen et al. \cite{method_GFCIL_zheng_2022} proposed GPIL, which implements lifelong learning in graphs to learn both newly encountered and previous classes. To enable a lifelong learning setting, firstly, the model spans the dataset into two parts, i.e., base classes and pseudo novel classes, with disjoint label spaces. Then, Graph FGL recurrently pre-trains the encoder on the base classes and keeps it during the pseudo incremental learning process. Then it performs meta-learning to learn initialization with more transferable meta-knowledge with a novel pseudo incremental learning, GPIL. During each meta-training episode, All few-shot classification tasks are sampled from pseudo-novel classes and base classes to imitate the incremental process in the evaluation process. By doing that, each meta-learning episode is able to learn a transferable model initialization for the incremental learning phase. In order to minimize the catastrophic forgetting problem, this method proposes a Hierarchical Attention Graph Meta-learning framework (HAG-Meta) by using the regularization technique to modify the loss function during the learning process. It uses a dynamically scaled loss regularizer that utilises the scale factor \cite{gfcil_scaleFactor_lin2017,gfcil_scaleFactor_menon2020l} to multiple task levels to adjust their contribution to model training. That scaling factor helps to reduce the contribution of insignificant classes while maximizing the important tasks. Moreover, This method proposed two hierarchical attention modules: 1) tasks-level attention, which estimates the importance of each task to balance the contribution of different tasks, and 2) node-level attention, which maintains a better balance between prior knowledge and new knowledge within the nodes. The model claims to have excellent stability in consolidating knowledge and gaining advantageous adaptability to new knowledge with minimal data samples.

The experiment conducted on GPIL uses a few-shot learning case study with graphs of few-shot learning datasets such as DBLP~\cite{tang2008arnetminer_dblbDataset}, Amazon-clothing~\cite{mcauley2015inferring_amazon}, and Reddit~\cite{bench_reddit_hamilton2017inductive}. It uses a split strategy for the class into 3 parts pre-training, meta-training, and evaluation. GPIL is compared with several methods such as continual learning on graphs, ER-GNN~\cite{method_ergnn_zhou_2020}, standard continual learning ICARL~\cite{method_icarl_rebuffi_2017}, which uses modifications with GNN, Few-shot Class-incremental Learning (CEC)~\cite{zhang2021few_CECmethod}, etc. The experiment calculates the performance dropping rate (PD) that measures the accuracy drop from the last sessions. There are 10 sessions in every experiment of each baseline method. GPIL achieve the lowest performance dropping rate (PD) in the DBLP dataset with 17.11\%. It is better than other baseline methods in terms of minimizing performance degradation.

\subsection{Topology-aware Weight Preserving}
The TWP method by Liu et al. \cite{method_TWP_liu_2020} proposed a novel method to strengthen lifelong learning and minimize catastrophic forgetting in GNNs. TWP explicitly studies the local structure of the input graph and stabilizes the important parameters in topological aggregation. Given an input graph and its embedding feature of nodes, TWP estimates the important score of each network parameter based on their contribution to the topological structure and task-related performance. The methods used in TWP are to calculate the gradients of the task-wise objective and topological preserving with each parameter, then consider the gradient as an index for the parameter importance.
After learning previous tasks, the model gets the optimal parameters by minimizing the loss on the task. As not all parameters contribute equally, it is important to preserve the minimized loss by considering highly influential parameters. Approximate contributions of each parameter are calculated based on an infinitesimal change in each parameter, like in the general lifelong learning method of Synaptic Intelligence (SI) \cite{method_SI_zenke2017}. Parameters that significantly contribute to minimizing loss must be kept stable when learning future tasks. Besides minimized loss preserving, topological structure preservation is conducted since structure information in the graph plays an important role. It aims to find the parameters with a substantial contribution to learning the topological information of the graph. GAT \cite{dicgrl_ne_GAN_velivckovic2017} are employed to learn topological information around the center node by calculating the attention coefficient of its neighbor. With the graph modeling process by GAT, then, the infinitesimal changes of each embedding feature are calculated to get the importance scores of each parameter. The regularization approach accommodates the new task by minimizing the performance degradation of the previous task. TWP penalizes changes in important parameters for old knowledge by regularizing the loss function while learning a new task. When changing high-importance parameters is penalized, the model can still perform well on previous tasks. The loss function for new tasks $\mathcal{T}_{k+1}$ formulated as:

\begin{equation}
\mathcal{L}_{k+1}^\prime (W) = \mathcal{L}^{new}_{k+1} + \sum^k_{n=1}I_n \otimes (W-W^*_n)^2,
\end{equation}
where $\mathcal{L}^{new}_{k+1}$ is a new loss function \emph{e.g.}, cross-entropy based on the task. $I_n$ denotes the network importance for the old tasks. $\otimes$ means element-wise multiplication and $W^*_n$ contains the optimal parameter for task $\mathcal{T}_n$. That learning strategy ensures that low-important score parameters can be adjusted freely while high-importance parameters will be maintained and penalized so that the model can handle the previous tasks. TWP also employs a strategy to promote the minimization of the computed importance score to maintain the model plasticity. 

One of the experiments on this method uses a node classification case study. Some datasets are used on node classifications, such as Corafull~\cite{bojchevski2017deep_corafulldataset}, amazon computer~\cite{mcauley2015inferring_amazon}, protein-protein interaction (PPI)~\cite{intro_drug_zitnik_2018} and Reddit~\cite{bench_reddit_hamilton2017inductive}. Three backbones of GNN are used to perform representation learning, such as GAT~\cite{dicgrl_ne_GAN_velivckovic2017}, GCN~\cite{motive_GCN_kipf_2016}, and GIN~\cite{xu2018powerful}, which are connected using several general lifelong learning such as LWF~\cite{method_lwf_li_2016}, GEM~\cite{method_GEM_lopez2017gradient}, EWC~\cite{method_ewc_kirkpatric_2016}, etc. TWP got the highest average performance with 0.976 and got the lowest average forgetting score of 0.001 $\pm$ 0.062 on the Reddit dataset. TWP demonstrates the effectiveness and ability to perform lifelong learning in the graph domain compared to other standard lifelong learning methods.

\subsection{Translation-based Knowledge Graph Embedding}
Song and Park \cite{method_translationBased_song_2018} developed a translation-based knowledge graph embedding method through continual learning that aims to enrich the representation of knowledge graphs when incoming representation appears in graph data. Moreover, this method minimizes the risk over new triples using rapid parameters that are penalized between the old and new embedding models. The translation-based knowledge graph embeddings aim to generate a vector representation in the embedding space of the knowledge graph based on the entity relation information in a graph. The simplest model in the translation-based knowledge graph embedding is TransE \cite{dicgrl_kgs_transe_bordes2013}, which forms the embedding matrix of entities and relations in the graph. In that case, the parameters to create the embedding are optimized using stochastic gradient descent.

The assumption is that the knowledge graph embedding method \cite{dicgrl_kgs_transe_bordes2013,dicgrl_kgs_rotate_sun2019G} is those graph entities whose relations are fixed after the training. This method employs the regularization term in order to accommodate the new embedding representation and preserves the representation of previously generated embedding. The underlying concept of the regularization approach in this model is to minimize empirical risk $\mathcal{R}^n_{emp}(\theta^n)$ defined as:

\begin{equation}
\mathcal{R}^n_{emp}(\theta^n) = \mathcal{R}^n_{emp}(D_n;\theta^n)+\lambda_n\phi(\theta^n,\theta^{n-1}),
\end{equation}
where $\mathcal{R}^n_{emp}(D_n;\theta^n$) denotes the empirical risk of new data. $\phi(\theta^n,\theta^{n-1})$ is as regularization term to penalize new parameter so that $\theta^n$ close to  $\theta^{n-1}$. $\lambda_n$ is a hyperparameter that maintains how much $\theta^n$ retrains the knowledge from previous data. As a result, the method can achieve the objective by minimizing risk penalized by rapid representation changes, and the old experiences are preserved in a new model.

One of the experiments with this method is in a knowledge graph embedding scenario with two knowledge graph datasets: WN18 (WordNet)~\cite{dettmers2018convolutional_WNdataset} and FB15k (Freebase)~\cite{toutanova2015observed_FBdataset}. New instances on nodes and relations are set to arrive sequentially. There are several baseline models used for comparison: 1) join learning as an upper bound, 2) finetuning as a lower bound that accommodates the new information continuously without any knowledge consolidation, and 3) general lifelong learning, EWC ~\cite{method_ewc_kirkpatric_2016}. The highest result of the experiment achieved 94.27 of H @10 (average top-10 ranked entities) score in the tripe classification task on the WN18 dataset. It is better than EWC and finetuning methods and almost equal to the upper bound joint learning technique.

\section{Hybrid Approach}
\label{hybrid}
The hybrid approach combines more than one lifelong learning approach to take advantage of each approach and maximize the performance of models \cite{method_GEM_lopez2017gradient,motive_categorization_maltoni_2019, method_GDM_parisi2018}. Examples of the graph lifelong learning models that implement hybrid approaches discussed in this section include ContinualGNN \cite{method_continualGNN_wang_2020}, LDANE \cite{method_LDANE_wei_2019}, and TrafficStream \cite{method_trafficStream_chen_2021}.

\subsection{ContinualGNN}
In the actual scenario model of ContinualGNN proposed by Wang et al. \cite{method_continualGNN_wang_2020}, there is a concept of replaying strategy to refine the network as a complement to the regularization approach itself. It combines both approaches to mitigate catastrophic forgetting and maintain the existing learned pattern. The first main goal of this model is to detect a new pattern in graph structure that significantly influences all nodes in the network. Second, the model aims to consolidate the knowledge in the entire network. The ContinualGNN model captures new patterns in streaming graph data. The model proposes a method based on a propagation process to efficiently mine the information on affected nodes when learning new patterns. The existing knowledge from the propagation process is then maintained using a combination of both approaches of rehearsal and regularization strategies. 

During the detection process for new patterns, ContinualGNN considers the response to the changes by looking at the number of changes in the networks. It is insufficient to consider new patterns when the changes are too small, for example, by only adding an edge in the network while the other neighborhood in the graph is very stable. Retraining that scenario is costly and a waste of time. When nodes' representation keeps evolving over time, the model will consider retraining the network with some strategies to make it efficient. A scoring function to calculate the influenced degree of a node that corresponds to the new patterns defined as: 

\begin{equation}
\mathcal{I}(\Delta G^t) = \{u\mid\parallel\Delta h^{t,L}_u \parallel  = \parallel  h^{t+\Delta t,L}_u - h^{t,L}_u \parallel > \delta\}.
\end{equation}

A higher score reflects that a node has a high influence. To calculate that new node representations, $h^{t+\Delta t,L}_u$ is compared with the previous representation version $h^{t,L}_u$ and results the influence scores. The model also uses threshold values $\delta$ to consider the number of nodes treated as a new pattern. Therefore, smaller threshold values will consider more nodes as new patterns only when the result of the influence score is above the thresholds and vice versa. The model limits to calculating the scoring function for all nodes at each time step because not all nodes are affected by the changes. Instead, the model only calculates it within the order $L$ neighbors that are affected. Furthermore, the model proposed two prospectives of data-view and model view, to consolidate the existing patterns. Data-view aims to maintain memory stability and select only important information of nodes and neighbors of some graph data to be saved and revisited during incremental training. ContinualGNN proposes a step-wise sampling strategy based on a reservoir sampling algorithm to achieve that objective \cite{method_continualGNNReservoir}. Model view, on the contrary, aims to solve the overfitting phenomenon by replaying a small amount of data in memory. This process is highly correlated to the regularization-based approach by using a generalization of knowledge preservation derived from a regular lifelong learning algorithm, EWC \cite{method_ewc_kirkpatric_2016}. This strategy maintains the distance between the current model parameters and the previous model parameters so that it helps to maintain the knowledge and performance of prior tasks in GNN.

Node classification case study is used to calculate the method's effectiveness. Some datasets, such as Cora~\cite{bench_coraCiteseerPubmed_sen2008}, Elliptic~\cite{weber2019anti_ellipticDataset}, and DBLP~\cite{tang2008arnetminer_dblbDataset} are used. Some baselines can be used, such as GNN with a retraining process as the upper bound and GNNs with incremental scenarios like finetuning as the lower bound. ContinualGNN achieves a maximum average accuracy of 0.9212 on the elliptic dataset. It proves to have better performance than the lower bound of GNNs with a simple incremental learning scenario and has almost similar performance to the upper bound methods of GNNs with a joint training strategy.

\subsection{Lifelong Dynamic Attributed Network Embedding}
Wei et al. \cite{method_LDANE_wei_2019} developed LDANE that aims to represent learning to produce a low-dimensional vector for each node in a growing size network. LDANE is developed based on lifelong learning that uses an architectural approach, namely Dynamic Expandable Networks (DEN) \cite{method_DEN_yoon2017}. DEN can dynamically determine network capacity to learn and share the structure among different tasks. Moreover, LDANE constructs attribute constraints that can be updated efficiently when the node attribute changes to restrict the learned embedding to satisfy the constraints. In this method, a Deep autoencoder is employed to learn the embedding of nodes. LDANE uses three kinds of loss functions of $L_{glob}$, $L_{loc}$, and $L_{attr}$. For any nodes $v_i$, $v_j$, $v_p$ and $v_m$, where $v_i$ is the observed node, $v_j$ is the neighbor node of $v_i$, $v_p$ is the most similar node-set in terms of attributes of node $v_i$, and $v_m$ is the most dissimilar node set of attributes of node $v_i$. $L_{glob}$ calculates the reconstruction error of output and input or error of global structure approximation that is shown as follows:

\begin{equation}
    L_{glob}=\sum^n_{i=1}\parallel(\hat{S_i}-S_i) \odot b_i\parallel^2_2.
\end{equation}
the weighted adjacency matrix of graph $G$ is denoted by $S$. if there is a relation from node $v_i$ and $v_j$, then $S_{ij}>0$; otherwise $S_{ij}=0$. $\hat{S_i}-S_i)$ defines a new reconstruction value of adjacency matrix and $\odot$ represents Hadamard product, $b_i$ denotes as a vector with $b_{ij}=1$ if $S_{ij}=0$; otherwise $b_{ij}=\delta>1$. It is not enough to preserve global structure proximity; this model also calculates $L_{loc}$ to preserve the local structure of embedding of its neighbor node $v_j$ that is defined as follows:

\begin{equation}
    L_{loc}=\sum^n_{i=1}S_{ij}\parallel(y_i-y_j)\parallel^2_2,
\end{equation}
where $y_i$ and $y_j$ denotes the result of node representation between observed node $v_i$ and $v_j$. LDANE also constructs two node sets $pos_i$ that contains top $l$ similar nodes $v_p\in pos_i$ and $neg_i$ that contains top $l$ dissimilar nodes $v_m\in neg_i$. The third loss function, $L_{attr}$ forces embedding to satisfy the constraints based on the top similar and dissimilar nodes:
\begin{equation}
L_{attr}=\sum_{v_i\in V}\sum_{v_m\in neg_i}\sum_{v_p\in pos_i} max(0,sim(y_i,y_m) - sim(y_i-y_p))
\end{equation}
where $sim(y_i,y_m)$ calculates cosine similarity between two embeddings of node $v_i$ and a dissimilar node $v_m$. $sim(y_i-y_p)$ calculate cosine similarity between two embeddings of $v_i$ and a similar node $v_p$. The main purpose of that loss function is to ensure that the similarity of similar node embeddings is not less than the similarity of dissimilar node embeddings. LDANE combines all objective functions of global structure proximity, local structure proximity, and attribute similarity together as follows:

\begin{equation}
    L=L_{glob} +\alpha_1 L_{loc} + \alpha_2 L_{attr} + v_1L_1 + v_2L_2,
\end{equation}
where $a_1$, $a_2$, $v_1$, $v_2$ are hyperparameters for relative weights of objective functions. $L_1$ and $L_2$ are regularizers to prevent the network weights from overfitting.

LDANE performs several processes, such as output expansion, selective retraining, autoencoder expansion, and autoencoder split, to enable lifelong learning on the network. Furthermore, it applies all approaches simultaneously. First, input and output expansion aims to accommodate a dynamic network that constantly changes. When the number of nodes at the following time observation is greater than the previous time observation, the LDANE architectural approach will increase the number of input and output neurons in the autoencoder to adjust the input data size. Then, selective retraining and expansion using the rehearsal principle will retrain the weights affected by network changes. Autoencoder split is conducted after the network embeddings are obtained if needed. It aims to evaluate the stability of network embedding over time by using the regularization method. The regularization approach is employed to regulate the parameter if it shifts too much from the original value and prevents forgetting. If the current features have significantly changed while the training process is running, then the network needs to train the weight again by splitting the autoencoder. 

A multilabel classification case study is one of the experiments that are used to calculate the performance of LDANE. Some datasets are used, such as synthetic data (SYN-1 and SYN-2)~\cite{benyahia2016dancer_syndataset}, DBLP~\cite{tang2008arnetminer_dblbDataset}, and Epinions\footnote{www.epinions.com}. Some comparison methods for embedding models include DeepWalk~\cite{perozzi2014deepwalk}, DynGEM~\cite{dynamic_dyngem_goyalDyngemDeepEmbedding2018}, and SDNE~\cite{wang2016structural_SDNE}. On multilabel classification results, LDANE achieved the highest average performance of micro-F1 0.93 and macro-F1 0.92 on the SYN-2 dataset. It performs better than other embedding network models that do not employ lifelong learning techniques.

\subsection{TrafficStream}
Chen et al. \cite{method_trafficStream_chen_2021} proposed a framework of streaming traffic flow forecasting called TrafficStream to address a specific domain of traffic flow that data structure is constantly evolving in its nodes and data structure. Apart from the ways to enable a lifelong learning setting, this type of method can be classified as a task-specific method. Task-specific methods usually have specific tasks that need to be achieved in order to get a good performance in the desired domain. The model (or method) is very focused on solving particular tasks and has no guarantee to be applicable to other domains. In general lifelong learning, examples of task-specific methods include NELL (never-ending language learner) \cite{motive_neverEnding_mitchell} and LSC (lifelong sentiment classification) \cite{method_lifelongSentiment_chen_2015}.

In order to address that challenge, the easiest way is to retrain the model from scratch regularly at a particular time, such as weekly, monthly, or yearly. However, it is considered very costly and inefficient for computing resources because it incorporates a massive amount of data when it comes sequentially. Moreover, because of network expansions, the topology and architecture of the previous ones will have different patterns, so the spatial dependency of previous models will not be relevant again. The previous insight from learned knowledge needs to be consolidated and preserved to help increase the model's performance. Based on those challenges, the TrafficStream framework aims to efficiently and effectively capture patterns for new graph structures and consolidate the knowledge between the previous and the new ones. In its framework, TrafficStream uses a simple traffic flow forecasting model (SurModel) to represent a complex graph data model in the intelligent transportation system \cite{method_stgcn_yu2017}. There are two component processes to efficiently update the current network pattern in the TrafficStream method: expanding network training and evaluation pattern detection. Those are based on the changes in network structures due to several conditions, such as adding new sensors or stations to record traffic flow data. The primary objective of expanding the training process is to reduce the complexity of capturing dependency between nodes in the graph. With vast amounts of attributes and a number of nodes, it will increase the processing task in the GNN layer. So that TrafficStream introduces the model to generate a sub-graph by reducing the number of included nodes to only 2-hops neighbors in a node. This mechanism claims to be effective in increasing the speed time of processing in GNN. The second component is evolution pattern detection, which aims to detect the different patterns on a graph entirely different from the previous one. This process is beneficial in recognizing the shifting of point of interest (POI) so that the model can selectively choose the nodes that have significant changes to be learned.
 
The essence of activating lifelong learning in this method is to consolidate knowledge to avoid catastrophic forgetting. For this reason, the model tracks back to the previous data and applies two strategies: information replay and smoothing parameters. Information replay performs the training process using sample data from previous information. The selection of sample data is simple by selecting randomly from several nodes. With the proposed JS divergence algorithm, the model chooses the nodes with lower JDS scores to be selected and used to construct sub-graphs. The second strategy is parameter smoothing, which aims to maintain shift information, resulting in catastrophic forgetting. The method used to perform parameter smoothing refers to the EWC method \cite{method_ewc_kirkpatric_2016} as defined below:

\begin{equation}
    \mathcal{L}_{smooth}=\lambda \sum_i F_i (\Psi_\mathcal{T}(i) - \Psi_{\mathcal{T}-1}(i))^2, 
\end{equation}
where $\lambda$ denotes the weight of the smoothing term, and $F_i$ refers to the importance of the $i$-th parameter in the previous function model $\Psi_\mathcal{T}(i)$ that can be estimated as follow:

\begin{equation}
    F = \frac{1}{| X_{\mathcal{T}-1} |} \sum_{x \in X_{\mathcal{T}-1}} [g(\Psi_{\mathcal{T}-1};x)g(\Psi_{\mathcal{T}-1};x)^T],
\end{equation}
where $g$ refers to first-order derivatives of the loss. $X_{\mathcal{T}_1}$ is a flow data of all nodes on previous graph $G_{\mathcal{T}-1}$. Through that mechanism, the weight with fewer essential parameters is smaller, and those parameters can adapt to the new patterns. On the other hand, more important parameters will have a larger weight in the weight-smoothing process so that changes can be limited to maintain previous knowledge.

The experiments are conducted to verify the model with traffic forecasting scenarios. It uses a real-world dataset called PEMS3-Stream~\cite{chen2001freeway_trafficstreamDataset}. Several traffic forecasting methods use a retraining strategy as a comparison baseline, STGCN~\cite{method_stgcn_yu2017}, spatial-temporal synchronous graph convolutional network (STSGCN)~\cite{song2020spatial_stsgcn}, gated recurrent unit (GRU)~\cite{cho2014learning_GRU}, etc. The conclusion shows that TrafficStream gets high prediction accuracy compared with the traditional retraining strategies, and also it is proven to reduce training complexity by using a continual learning strategy in graph-based tasks.

\section{Method Comparisons}
\label{'sec:method_compare'}
Comparing current methods of graph lifelong learning still remains an open issue. Although graph lifelong learning focuses on accommodating new knowledge, preventing catastrophic forgetting, and considering graph representation learning, current methods have various scenarios and applications. Those scenarios can be grouped into three different types: 1) only focusing on new instances (NI), (2) new classes (NC), or (3) both new instances and new classes (NIC). Moreover, each model has different applications, such as node prediction, link prediction, node embedding, and others.

This section summarises the considerations for each categorization of lifelong learning. For the architectural approach, The models adjust their complexity by allocating prototypes like HPNs \cite{method_HPNs_zhang_2021}, and expansions in LDANE \cite{method_LDANE_wei_2019}. The model should consider memory allocations and efficiency aspects with those expansion mechanisms to prevent a technical bottleneck. In the regularization approach, the way to allocate new knowledge is by managing the shared solution space to be applicable to all tasks. However, managing the constraints properly on model weight should be considered because new tasks are forced to adapt to the previous knowledge, resulting in no maximum performance to address new tasks. For the rehearsal approach, some samples from prior experiences are replayed to strengthen memory and maintain performance to complete previous tasks. Using maximum data resources should be avoided because it will involve very complex data and attributes in the graph. Therefore, considering computational speed is very important when retraining past knowledge so that the model does not require extensive time complexity. The strategy of selecting nice and small samples of data, nodes, and topological structures in graph data involved through the replay process should be well considered to minimize the time complexity of the learning process. An automated algorithm should be able to select the nice samples automatically to enhance the quality of the retraining process \cite{qu2020generalOptimize}. Moreover, a strategy to determine when to activate the replay process is also needed not to run it every time to be more efficient. For the hybrid method, even though it will take full advantage of specific approaches of architecture, regularization, and rehearsal and has a more complex framework, it needs to address multiple considerations across different approaches. All models that have been reviewed have their unique characteristics based on the scenario and its applications. Based on those unique characteristics, the summary of the models' differences can be seen in Table~\ref{tb:methodscenarios}.

\begin{table*}
\centering
\caption{Method Comparisons}
\label{tb:methodscenarios}
\resizebox{\textwidth}{!}{
\begin{tabular}{|l|l|l|l|l|} \hline
\multicolumn{1}{|c|}{Methods} & \multicolumn{1}{c|}{Applications} & \multicolumn{1}{c|}{Graph Learning
  Modules} & \multicolumn{1}{c|}{Scenarios} & \multicolumn{1}{c|}{Metrics} \\ \hline
\begin{tabular}[c]{@{}l@{}}FGN \cite{method_FGN_wang_2020}\end{tabular} & classification & \begin{tabular}[c]{@{}l@{}}graph convolution, \\graph attention\end{tabular} & \begin{tabular}[c]{@{}l@{}}new instances \&\\new classes (NIC)\end{tabular} & \begin{tabular}[c]{@{}l@{}}average accuracy,\\forgetting measure\end{tabular} \\ \hline
\begin{tabular}[c]{@{}l@{}}HPNs \cite{method_HPNs_zhang_2021}\end{tabular} & classification & atomic structure
  embedding & \begin{tabular}[c]{@{}l@{}}new instances \&\\new classes (NIC)\end{tabular} & \begin{tabular}[c]{@{}l@{}}average accuracy,\\forgetting measure\end{tabular} \\ \hline
\begin{tabular}[c]{@{}l@{}}ER-GNN \cite{method_ergnn_zhou_2020}\end{tabular} & classification & graph neural network & \begin{tabular}[c]{@{}l@{}}new instances \&\\new classes (NIC)\end{tabular} & \begin{tabular}[c]{@{}l@{}}average accuracy,\\forgetting measure\end{tabular} \\ \hline
\begin{tabular}[c]{@{}l@{}}LONC\cite{method_openWorld_2021}\end{tabular} & classification & graph neural network & new classes (NC) & \begin{tabular}[c]{@{}l@{}}average accuracy,\\forward transfer\end{tabular} \\ \hline
\begin{tabular}[c]{@{}l@{}}DiCGR \cite{method_DiCGRL_kou_2020}\end{tabular} & \begin{tabular}[c]{@{}l@{}}link prediction, \\node classification\end{tabular} & \begin{tabular}[c]{@{}l@{}}knowledge graph embedding, \\graph attention\end{tabular} & new instances (NI) & \begin{tabular}[c]{@{}l@{}}average top-10 \\ranked entities \\(H@ 10) \&\\ average accuracy\end{tabular}\\ \hline
\begin{tabular}[c]{@{}l@{}}GPIL \cite{method_GFCIL_zheng_2022}\end{tabular} & few-shot node classification & graph convolution & \begin{tabular}[c]{@{}l@{}}new instances \&\\new classes (NIC)\end{tabular} & \begin{tabular}[c]{@{}l@{}}average accuracy,\\dropping rate\end{tabular} \\ \hline
\begin{tabular}[c]{@{}l@{}}TWP \cite{method_TWP_liu_2020}\end{tabular} & node and graph
  classification & graph neural network & new classes (NC) & \begin{tabular}[c]{@{}l@{}}average accuracy,\\forgetting measure\end{tabular} \\ \hline
\begin{tabular}[c]{@{}l@{}}TKGE \cite{method_translationBased_song_2018}\end{tabular} & \begin{tabular}[c]{@{}l@{}}knowledge graph enrichment, \\link prediction, triple classification\end{tabular} & knowledge graph
  embedding & new instances (NI) & \begin{tabular}[c]{@{}l@{}}average top-10 \\ranked entities \\(H@ 10)\end{tabular} \\ \hline
ContinualGNN
  \cite{method_continualGNN_wang_2020} & classification & graphSAGE & \begin{tabular}[c]{@{}l@{}}new instances \&\\new classes (NIC)\end{tabular} & average accuracy \\ \hline
\begin{tabular}[c]{@{}l@{}}LDANE \cite{method_LDANE_wei_2019}\end{tabular} & \begin{tabular}[c]{@{}l@{}}network representation enrichment, \\graph reconstruction, link prediction, \\node classification\end{tabular} & network embedding & new instances (NI) & average precision \\ \hline
TrafficStream
  \cite{method_trafficStream_chen_2021} & traffic flow
  forecasting & graph neural network & new instances (NI) & mean errors \\ \hline
\end{tabular}
}
\end{table*}

\section{Benchmarks}
\label{sec:benchmarks}
This section briefly explains some of the available datasets, experimental details, and how to calculate performance on graph lifelong learning.
\subsection{Datasets}
There are no datasets that are specifically intended to check the performance of graph lifelong learning. Most of the benchmark datasets used in general lifelong learning are in the form of image classification tasks, such as Permuted MNIST \cite{method_ewc_kirkpatric_2016} and Split CIFAR \cite{method_GEM_lopez2017gradient}. Those datasets can still be used as graph representations, such as changing “superpixels” to represent nodes in the graph \cite{benchmark_carta2021catastrophic}. However, assuming superpixels to represent nodes in the graph is not close to the real scenario of graph structure that forms an irregular structure that differs from a grid structure in the image.

Several popular graph learning benchmark datasets such as citation datasets (Citeseer \cite{bench_coraCiteseerPubmed_sen2008}, Cora\cite{bench_coraCiteseerPubmed_sen2008}, OGBN-Arxiv \cite{bench_arxiv_mikolov2013distributed, bench_arxiv_wang2020microsoft}, and Pubmed \cite{bench_coraCiteseerPubmed_sen2008}), User comment datasets (Reddit \cite{bench_reddit_hamilton2017inductive}), product co-purchasing network (OGBN-product \cite{bench_products_chiang2019cluster}), and protein interaction network (OGBN-Proteins \cite{bench_proteins_szklarczyk2019string}) can be used to test the performance of the graph lifelong learning through scenario modification. The modification of new instances and new classes scenario is by splitting the class or label of the tasks to be distributed at several different times. For example, there are 10 prediction classes needed to enable an experiment on graph lifelong learning, those classes need to be split into several incremental learning scenarios, for instance, five tasks, and it will result in two classes per task. The process starts with learning the first task and incrementally learning the rest of the tasks by sequentially adding new instances and new classes. This experimental design aims to determine whether performance methods can perform incremental learning, accommodate new tasks, and prevent catastrophic forgetting of previous tasks.

\begin{table}[]
\centering
\caption{Datasets}
\resizebox{\columnwidth}{!}{%
\begin{tabular}{|l|l|l|l|l|}
\hline
\multicolumn{1}{|c|}{\textbf{Dataset}} & \multicolumn{1}{c|}{\textbf{Nodes}} & \multicolumn{1}{c|}{\textbf{Edges}} & \multicolumn{1}{c|}{\textbf{Features}} & \multicolumn{1}{c|}{\textbf{Tasks}} \\ \hline
Citeseer & 3,312 & 4,732 & 3,703 & 6 \\ \hline
Cora & 2,708 & 5,429 & 1,433 & 7 \\ \hline
OGBN-Arxiv & 169,343 & 1,166,243 & 128 & 40 \\ \hline
OGBN-Product & 2,449,029 & 61,859,140 & 100 & 47 \\ \hline
OGBN-Proteins & 132,534 & 39,561,252 & 8 & 112 \\ \hline
Pubmed & 19,717 & 44,338 & 500 & 3 \\ \hline
Reddit & 232,965 & 11,606,919 & 602 & 41 \\ \hline
\end{tabular}%
}
\end{table}

\subsection{Performance Metrics}
Specifying evaluation metrics before measuring the performance of graph lifelong learning methods is important. Evaluation metrics in lifelong learning graphs are not similar to calculating the accuracy of tasks in general machine learning or graph learning. Chaudhry et al. \cite{metrics_first_chaudhry2018riemannian} explained the most common metric to measure performance in general lifelong learning, such as average accuracy (A) and forgetting measure (F). Another proposed work by Chaudhry et al. \cite{metrics_second_chaudhry2018efficient} described a new measure called the learning curve area (LCA) that calculates how fast a model learns. The details of those three metrics are as follow:

\begin{itemize}
   \item \textbf{Average Accuracy ($A \in [0,1]$)} \cite{metrics_first_chaudhry2018riemannian}. The average accuracy of all tasks after processing incremental learning from the first task to the $T$-th task is defined as:
   
\begin{equation}
A_T =\frac{1}{T}\sum_{i=1}^{T}a_{T,i}
\end{equation}

where $a_{T,i} \in [0,1]$ is the accuracy of individual task on $i$-th task ($i \leq T$) that calculated after learning incrementally from task 1 to task $T$.

   \item \textbf{Forgetting Measure ($F \in [-1,1]$)} \cite{metrics_first_chaudhry2018riemannian}. The average forgetting measure of all tasks after incremental process learning from the first task to the T task is defined as:

\begin{equation}
F_T=\frac{1}{T-1}\sum_{i=1}^{T}f_{i}^{T}
\end{equation}

where $f_{i}^{j}$ is the forgetting value on individual task $t_i$ after the model learn up to task $t_j$ and it is defined as:

\begin{equation}
f_{i}^{j}=\max_{k\epsilon \left \{ 1,...,j-1 \right \}}a_{k,i}-a_{j,i}
\end{equation}

   \item \textbf{Learning Curve Area ($LCA \in [0,1]$)} \cite{metrics_second_chaudhry2018efficient}. LCA is the area under $Z_b$ curve that calculates average $b$-shot performance (where $b$ denotes a mini-batch number). $Z_b$ is the accuracy of each task after observing $ b$-th mini-batch. It will be high if the zero-shot performance is good and a model learns quickly:
\begin{equation}
Z_b=\frac{1}{T}\sum_{i=1}^{T}a_{i,b,i}
\end{equation}

\end{itemize}

\section{Open Issues}
\label{sec:issue}
The biggest challenge in lifelong learning settings is catastrophic forgetting~\cite{motive_catastrophicForgetting_french_1999} or called catastrophic interference \cite{openIssue_catastrophicInterference_gordon}, where a model cannot maintain performance on previous tasks when learning new tasks, as shown in Fig. \ref{forgetting}. In graph lifelong learning, consolidating knowledge is necessary to minimize forgetting problems based on changes in the graph representation in terms of nodes, relations, and tasks. Examples of classical approaches that cannot optimally solve catastrophic forgetting in graph lifelong learning include finetuning and joint training \cite{contribute_delange_2021}. Finetuning uses a model from previous tasks to optimize the task's parameters being learned. When the model is not guided to learn new tasks, it results in performance degradation for completing previous tasks. Joint training is an approach that uses data from previous tasks and new tasks to conduct training together. It may provide optimal results but requires all the data that might have been lost, resulting in process inefficiency when having massive data so that learning iteratively with a small amount of data is needed.

In addition to catastrophic forgetting, there are some other open issues and challenges to implementing graph lifelong learning, as summarized in the following points:

\subsection{Uncertain Neighboorhoods}
In learning the representation of a graph, every node relationship should be converted into a lower-dimensional vector space so that it can be further processed in solving downstream tasks \cite{openissue_networkEmbedding_cai2018comprehensive}. The learning representation process involves changing node attributes' information with its neighbors. Referring to predefined graphs with fixed structures, feature embedding on nodes will be easier to conduct, so that neighborhood information is more stable for the learning process as is done in conventional GNNs methods~\cite{motive_GCN_kipf_2016, bench_reddit_hamilton2017inductive}. However, in graph lifelong learning, although some approach explicitly addresses the problem of the continual embedding process \cite{method_DiCGRL_kou_2020, method_LDANE_wei_2019, method_translationBased_song_2018}, it still requires an optimal strategy for dynamic embedding on graphs with uncertain neighborhood conditions. Graph lifelong learning should be able to capture new relations and nodes and consider when the right time to do the embedding process to accommodate data instances in order to improve the algorithms' scalability and efficiency.

\subsection{Extreme Evolution}
The evolution of graph data is complicated to understand. Sometimes the data graph turns out to be more complex but still retains most of the old structure. On the other hand, there are types of graph networks with extreme evolution, which means the graph structure can drastically evolve over time, and most of the old forms are extinct from the graph data. Moreover, when the graph data has an extreme evolution characteristic, it is likely that the previously learned tasks no longer need to be maintained. This factor in the graph also needs to be considered when implementing a graph lifelong learning. The model expects to judge whether it needs to minimize forgetting the previous tasks or let the model forget the previous tasks completely that are no longer relevant to the currently observed graph representation.

\subsection{Global Dependencies Learning}
Learning global context or global dependencies in graph data is very important to understand the entire relationships between nodes. It can capture the information of two nodes even if they stay apart in a large-scale graph. Current progress in learning global dependency optimally in a static graph has achieved state-of-the-art performances, for example, using transformer mechanisms that elevated sequential learning with novel positional encoding implemented in graph data~\cite{min2022transformer_graphTrans}. In graph lifelong learning, passing the information of new instances across all the nodes and edges in learning the global context of graphs incrementally is still challenging. Further improvement is needed to learn global dependency efficiently in a continual manner, especially in a large-scale graph, to improve graph lifelong learning methods.

\subsection{Comparative Benchmarking}
Current graph lifelong learning strategies are applied in different scenarios, such as focusing on adding new instances (NI), new classes (NC), or considering both new instances and classes (NIC). Moreover, it has different applications, such as node classification, graph prediction, edge prediction, or graph embedding enrichment. Nowadays, benchmarking strategies are highly non-standard. It is essential to be considered as further research in addition to the current studies to bring fair comparations and improve graph lifelong learning methods.

\subsection{Model Training Parallelisation}
Most graph lifelong learning techniques are run on single-point computational resources. However, when looking at the current practical implementation, most of the data comes from a distributed architecture that is deployed on spread-edge devices using cloud services. Collecting all data on a single point computational resource sometimes brings other limitations on security, performance, and costs. To address the emergence of more intelligent services in a distributed architecture, another open challenge of graph lifelong learning is to enable a federated learning paradigm \cite{zhang2021surveyFederated,zhang2021federated} that can perform computation in edge devices using local models parallelly and aggregate them to the single-point device into a general model \cite{usmanova2021distillation}. That concept will be beneficial because representations of nodes or relations of graphs and new classes can appear in local points of distributed architecture.

\section{Conclusion}
\label{sec:conclu}
This survey paper overviews graph lifelong learning, an emerging field at the intersection of graph learning and lifelong learning. The motivations to enable lifelong learning settings on graph data and several potential applications of graph lifelong learning in several fields, such as social networks, traffic prediction, recommender systems, and anomaly detection, are discussed in depth. This paper also covers the state-of-the-art methods in graph lifelong learning, which are categorized into several groups: architectural approach, rehearsal approach, regularization approach, and hybrid approach. It is also presented methods comparisons by considering some computational aspects in each categorization. All methods are compared based on each scenario and application. Moreover, explanation of benchmarking settings and candidate datasets for testing graph lifelong learning models are also provided. Apart from the catastrophic forgetting problem, several open issues of graph lifelong learning, such as uncertain neighborhoods, extreme evolution, global dependencies learning, fair benchmarking, class imbalance, and model training parallelization, are discussed as future research directions. It is worth noting that many common challenges facing general machine learning are also applicable to graph lifelong learning, including, \emph{e.g.}, fairness and bias, explainability/interpretability, privacy and security, and data trust/quality. Overall, our survey work facilitates new interests and works in the design and development of innovative theory, models/algorithms, and applications of graph lifelong learning.

\balance
\bibliographystyle{IEEEtran}
\bibliography{main}

\begin{thebibliography}{100}
\providecommand{\url}[1]{#1}
\csname url@samestyle\endcsname
\providecommand{\newblock}{\relax}
\providecommand{\bibinfo}[2]{#2}
\providecommand{\BIBentrySTDinterwordspacing}{\spaceskip=0pt\relax}
\providecommand{\BIBentryALTinterwordstretchfactor}{4}
\providecommand{\BIBentryALTinterwordspacing}{\spaceskip=\fontdimen2\font plus
\BIBentryALTinterwordstretchfactor\fontdimen3\font minus
  \fontdimen4\font\relax}
\providecommand{\BIBforeignlanguage}[2]{{%
\expandafter\ifx\csname l@#1\endcsname\relax
\typeout{** WARNING: IEEEtran.bst: No hyphenation pattern has been}%
\typeout{** loaded for the language `#1'. Using the pattern for}%
\typeout{** the default language instead.}%
\else
\language=\csname l@#1\endcsname
\fi
#2}}
\providecommand{\BIBdecl}{\relax}
\BIBdecl

\bibitem{contribute_wu_GNNSurvey_2019}
Z.~Wu, S.~Pan, F.~Chen, G.~Long, C.~Zhang, and S.~Y. Philip, ``A comprehensive
  survey on graph neural networks,'' \emph{IEEE transactions on neural networks
  and learning systems}, vol.~32, no.~1, pp. 4--24, 2020.

\bibitem{contribute_zhang_graphSurvey_2020}
Z.~Zhang, P.~Cui, and W.~Zhu, ``Deep learning on graphs: A survey,'' \emph{IEEE
  Transactions on Knowledge and Data Engineering}, vol.~34, no.~1, pp.
  249--270, 2022.

\bibitem{contribute_feng_graphSurvey_2021}
F.~Xia, K.~Sun, S.~Yu, A.~Aziz, L.~Wan, S.~Pan, and H.~Liu, ``Graph learning: A
  survey,'' \emph{IEEE Transactions on Artificial Intelligence}, vol.~2, no.~2,
  pp. 109--127, 2021.

\bibitem{intro_drug_zitnik_2018}
M.~Zitnik, M.~Agrawal, and J.~Leskovec, ``Modeling polypharmacy side effects
  with graph convolutional networks,'' \emph{Bioinformatics}, vol.~34, no.~13,
  pp. i457--i466, 2018.

\bibitem{intro_kg_wang_2018}
Z.~Wang, T.~Chen, J.~Ren, W.~Yu, H.~Cheng, and L.~Lin, ``Deep reasoning with
  knowledge graph for social relationship understanding,'' \emph{arXiv preprint
  arXiv:1807.00504}, 2018.

\bibitem{intro_socialnetwork_liu_2019}
J.~Liu, F.~Xia, L.~Wang, B.~Xu, X.~Kong, H.~Tong, and I.~King, ``Shifu2: A
  network representation learning based model for advisor-advisee relationship
  mining,'' \emph{IEEE Transactions on Knowledge and Data Engineering}, p.
  1–1, 2019.

\bibitem{chen2021heterogeneousRecommender}
X.~Chen, T.~Tang, J.~Ren, I.~Lee, H.~Chen, and F.~Xia, ``Heterogeneous graph
  learning for explainable recommendation over academic networks,'' in
  \emph{IEEE/WIC/ACM International Conference on Web Intelligence and
  Intelligent Agent Technology}, 2021, pp. 29--36.

\bibitem{kong2022exploringHumanMobility}
X.~Kong, K.~Wang, M.~Hou, F.~Xia, G.~Karmakar, and J.~Li, ``Exploring human
  mobility for multi-pattern passenger prediction: A graph learning
  framework,'' \emph{IEEE Transactions on Intelligent Transportation Systems},
  2022.

\bibitem{method_ergnn_zhou_2020}
F.~Zhou and C.~Cao, ``Overcoming catastrophic forgetting in graph neural
  networks with experience replay,'' in \emph{Proceedings of the AAAI
  Conference on Artificial Intelligence}, vol.~35, no.~5, 2021, pp. 4714--4722.

\bibitem{method_continualGNN_wang_2020}
J.~Wang, G.~Song, Y.~Wu, and L.~Wang, ``Streaming graph neural networks via
  continual learning,'' in \emph{Proceedings of the 29th ACM International
  Conference on Information \& Knowledge Management}, 2020, pp. 1515--1524.

\bibitem{lesort_lomonaco_2019_continual_survey}
T.~Lesort, V.~Lomonaco, A.~Stoian, D.~Maltoni, D.~Filliat, and
  N.~D{\'\i}az-Rodr{\'\i}guez, ``Continual learning for robotics: Definition,
  framework, learning strategies, opportunities and challenges,''
  \emph{Information fusion}, vol.~58, pp. 52--68, 2020.

\bibitem{metrics_first_chaudhry2018riemannian}
A.~Chaudhry, P.~K. Dokania, T.~Ajanthan, and P.~H. Torr, ``Riemannian walk for
  incremental learning: Understanding forgetting and intransigence,'' in
  \emph{Proceedings of the European Conference on Computer Vision (ECCV)},
  2018, pp. 532--547.

\bibitem{motive_neverEnding_mitchell}
T.~Mitchell, W.~Cohen, E.~Hruschka, P.~Talukdar, B.~Yang, J.~Betteridge,
  A.~Carlson, B.~Dalvi, M.~Gardner, B.~Kisiel \emph{et~al.}, ``Never-ending
  learning,'' \emph{Communications of the ACM}, vol.~61, no.~5, pp. 103--115,
  2018.

\bibitem{chen_liu_2018_LifelongMachineLearning}
Z.~Chen and B.~Liu, \emph{Lifelong machine learning}.\hskip 1em plus 0.5em
  minus 0.4em\relax Morgan \& Claypool Publishers, 2018, vol.~12, no.~3.

\bibitem{motive_definitionLifelong_ruvolo_ELLA}
\BIBentryALTinterwordspacing
P.~Ruvolo and E.~Eaton, ``{ELLA}: An efficient lifelong learning algorithm,''
  in \emph{Proceedings of the 30th International Conference on Machine
  Learning}, ser. Proceedings of Machine Learning Research, S.~Dasgupta and
  D.~McAllester, Eds., vol.~28, no.~1.\hskip 1em plus 0.5em minus 0.4em\relax
  Atlanta, Georgia, USA: PMLR, 17--19 Jun 2013, pp. 507--515. [Online].
  Available: \url{https://proceedings.mlr.press/v28/ruvolo13.html}
\BIBentrySTDinterwordspacing

\bibitem{motive_lifelongDefinition_parisi_2019}
G.~I. Parisi, R.~Kemker, J.~L. Part, C.~Kanan, and S.~Wermter, ``Continual
  lifelong learning with neural networks: A review,'' \emph{Neural Networks},
  vol. 113, pp. 54--71, 2019.

\bibitem{motive_catastrophicForgetting_french_1999}
R.~French, ``Catastrophic forgetting in connectionist networks,'' \emph{Trends
  in Cognitive Sciences}, vol.~3, no.~4, pp. 128--135, 1999.

\bibitem{motive_catastrophicForgetting_robins_1995}
A.~Robins, ``Catastrophic forgetting, rehearsal and pseudorehearsal,''
  \emph{Connection Science}, vol.~7, no.~2, pp. 123--146, 1995.

\bibitem{method_ewc_kirkpatric_2016}
J.~Kirkpatrick, R.~Pascanu, N.~Rabinowitz, J.~Veness, G.~Desjardins, A.~A.
  Rusu, K.~Milan, J.~Quan, T.~Ramalho, A.~Grabska-Barwinska \emph{et~al.},
  ``Overcoming catastrophic forgetting in neural networks,'' \emph{Proceedings
  of the national academy of sciences}, vol. 114, no.~13, pp. 3521--3526, 2017.

\bibitem{method_lwf_li_2016}
Z.~Li and D.~Hoiem, ``Learning without forgetting,'' \emph{IEEE transactions on
  pattern analysis and machine intelligence}, vol.~40, no.~12, pp. 2935--2947,
  2017.

\bibitem{method_progNN_rusu_2016_}
A.~A. Rusu, N.~C. Rabinowitz, G.~Desjardins, H.~Soyer, J.~Kirkpatrick,
  K.~Kavukcuoglu, R.~Pascanu, and R.~Hadsell, ``Progressive neural networks,''
  \emph{arXiv preprint arXiv:1606.04671}, 2016.

\bibitem{method_cwr_lomonaco_2017}
V.~Lomonaco and D.~Maltoni, ``Core50: a new dataset and benchmark for
  continuous object recognition,'' in \emph{Conference on Robot
  Learning}.\hskip 1em plus 0.5em minus 0.4em\relax PMLR, 2017, pp. 17--26.

\bibitem{method_GEM_lopez2017gradient}
D.~Lopez-Paz and M.~Ranzato, ``Gradient episodic memory for continual
  learning,'' \emph{Advances in neural information processing systems},
  vol.~30, 2017.

\bibitem{motive_categorization_maltoni_2019}
D.~Maltoni and V.~Lomonaco, ``Continuous learning in single-incremental-task
  scenarios,'' \emph{Neural Networks}, vol. 116, pp. 56--73, 2019.

\bibitem{method_DEN_yoon2017}
J.~Yoon, E.~Yang, J.~Lee, and S.~J. Hwang, ``Lifelong learning with dynamically
  expandable networks,'' \emph{arXiv preprint arXiv:1708.01547}, 2017.

\bibitem{openissue_networkEmbedding_cai2018comprehensive}
H.~Cai, V.~W. Zheng, and K.~C.-C. Chang, ``A comprehensive survey of graph
  embedding: Problems, techniques, and applications,'' \emph{IEEE Transactions
  on Knowledge and Data Engineering}, vol.~30, no.~9, pp. 1616--1637, 2018.

\bibitem{motive_GCN_kipf_2016}
T.~N. Kipf and M.~Welling, ``Semi-supervised classification with graph
  convolutional networks,'' \emph{arXiv preprint arXiv:1609.02907}, 2016.

\bibitem{motiveNodeliLearningDeepNeural2019}
B.~Li and D.~Pi, ``Learning deep neural networks for node classification,''
  \emph{Expert Systems with Applications}, vol. 137, pp. 324--334, 2019.

\bibitem{motiveLinkcaiLineGraphNeural2021}
L.~Cai, J.~Li, J.~Wang, and S.~Ji, ``Line {{Graph Neural Networks}} for {{Link
  Prediction}},'' \emph{IEEE Transactions on Pattern Analysis and Machine
  Intelligence}, pp. 1--1, 2021.

\bibitem{motiveGraphzhangEndtoendDeepLearning2018}
M.~Zhang, Z.~Cui, M.~Neumann, and Y.~Chen, ``An end-to-end deep learning
  architecture for graph classification,'' in \emph{Thirty-Second {{AAAI}}
  Conference on Artificial Intelligence}, 2018.

\bibitem{motiv_gnnModel_franco_2009}
F.~Scarselli, M.~Gori, A.~C. Tsoi, M.~Hagenbuchner, and G.~Monfardini, ``The
  graph neural network model,'' \emph{IEEE transactions on neural networks},
  vol.~20, no.~1, pp. 61--80, 2008.

\bibitem{motive_graphRNN_you_2018}
\BIBentryALTinterwordspacing
J.~You, R.~Ying, X.~Ren, W.~L. Hamilton, and J.~Leskovec, ``Graphrnn:
  Generating realistic graphs with deep auto-regressive models,''
  \emph{International Conference on Machine Learning}, Jun 2018. [Online].
  Available: \url{https://arxiv.org/abs/1802.08773v3}
\BIBentrySTDinterwordspacing

\bibitem{motive_GAE_kipf_2016b}
T.~N. Kipf and M.~Welling, ``Variational graph auto-encoders,'' \emph{arXiv
  preprint arXiv:1611.07308}, 2016.

\bibitem{motive_spatialTemporal_wu_2019}
Z.~Wu, S.~Pan, G.~Long, J.~Jiang, and C.~Zhang, ``Graph wavenet for deep
  spatial-temporal graph modeling,'' ser. IJCAI'19.\hskip 1em plus 0.5em minus
  0.4em\relax AAAI Press, 2019, pp. 1907--1913.

\bibitem{bench_reddit_hamilton2017inductive}
W.~Hamilton, Z.~Ying, and J.~Leskovec, ``Inductive representation learning on
  large graphs,'' \emph{Advances in neural information processing systems},
  vol.~30, 2017.

\bibitem{method_openWorld_2021}
L.~Galke, B.~Franke, T.~Zielke, and A.~Scherp, ``Lifelong learning of graph
  neural networks for open-world node classification,'' in \emph{2021
  International Joint Conference on Neural Networks (IJCNN)}.\hskip 1em plus
  0.5em minus 0.4em\relax IEEE, 2021, pp. 1--8.

\bibitem{relatedSurveykazemiRepresentationLearningDynamic2020}
S.~M. Kazemi, R.~Goel, K.~Jain, I.~Kobyzev, A.~Sethi, P.~Forsyth, and
  P.~Poupart, ``Representation learning for dynamic graphs: A survey.''
  \emph{J. Mach. Learn. Res.}, vol.~21, no.~70, pp. 1--73, 2020.

\bibitem{relatedSurveybarrosSurveyEmbeddingDynamic2021}
C.~D.~T. Barros, M.~R.~F. Mendon{\c c}a, A.~B. Vieira, and A.~Ziviani, ``A
  {{Survey}} on {{Embedding Dynamic Graphs}},'' \emph{ACM Comput. Surv.},
  vol.~55, no.~1, pp. 10:1--10:37, Nov. 2021.

\bibitem{overviewDynamicxueDynamicNetworkEmbedding2022}
G.~Xue, M.~Zhong, J.~Li, J.~Chen, C.~Zhai, and R.~Kong, ``Dynamic network
  embedding survey,'' \emph{Neurocomputing}, vol. 472, pp. 212--223, 2022.

\bibitem{dynamic_dyngem_goyalDyngemDeepEmbedding2018}
P.~Goyal, N.~Kamra, X.~He, and Y.~Liu, ``Dyngem: {{Deep}} embedding method for
  dynamic graphs,'' \emph{arXiv preprint arXiv:1805.11273}, 2018.

\bibitem{dynamic_dyngraph2vec_GOYAL}
P.~Goyal, S.~R. Chhetri, and A.~Canedo, ``dyngraph2vec: Capturing network
  dynamics using dynamic graph representation learning,'' \emph{Knowledge-Based
  Systems}, vol. 187, p. 104816, 2020.

\bibitem{dynamic_triad_zhouDynamic2018}
L.~Zhou, Y.~Yang, X.~Ren, F.~Wu, and Y.~Zhuang, ``Dynamic network embedding by
  modeling triadic closure process,'' in \emph{Proceedings of the {{AAAI}}
  Conference on Artificial Intelligence}, vol.~32, 2018.

\bibitem{applic_continualSocialNet_yi_2020}
Y.~Han, S.~Karunasekera, and C.~Leckie, ``Graph neural networks with continual
  learning for fake news detection from social media,'' \emph{arXiv preprint
  arXiv:2007.03316}, 2020.

\bibitem{method_trafficStream_chen_2021}
X.~Chen, J.~Wang, and K.~Xie, ``Trafficstream: A streaming traffic flow
  forecasting framework based on graph neural networks and continual
  learning,'' \emph{Proceedings of the Thirtieth International Joint Conference
  on Artificial Intelligence}, 2021.

\bibitem{appli_recommenderSystem_guo_2019}
T.~Guo, F.~Xia, S.~Zhen, X.~Bai, D.~Zhang, Z.~Liu, and J.~Tang, ``Graduate
  employment prediction with bias,'' in \emph{Proceedings of the AAAI
  Conference on Artificial Intelligence}, vol.~34, no.~01, 2020, pp. 670--677.

\bibitem{appli_anomaly_chaudhar_2019}
A.~Chaudhary, H.~Mittal, and A.~Arora, ``Anomaly detection using graph neural
  networks,'' in \emph{2019 International Conference on Machine Learning, Big
  Data, Cloud and Parallel Computing (COMITCon)}, 2019, pp. 346--350.

\bibitem{contribute_delange_2021}
M.~Delange, R.~Aljundi, M.~Masana, S.~Parisot, X.~Jia, A.~Leonardis,
  G.~Slabaugh, and T.~Tuytelaars, ``A continual learning survey: Defying
  forgetting in classification tasks,'' \emph{IEEE Transactions on Pattern
  Analysis and Machine Intelligence}, p. 1–1, 2021.

\bibitem{motive_categorization_biesialska_2020_survey}
M.~Biesialska, K.~Biesialska, and M.~R. Costa-Jussà, ``Continual lifelong
  learning in natural language processing: A survey,'' \emph{Proceedings of the
  28th International Conference on Computational Linguistics}, 2020.

\bibitem{hung2019compacting_CPG}
C.-Y. Hung, C.-H. Tu, C.-E. Wu, C.-H. Chen, Y.-M. Chan, and C.-S. Chen,
  ``Compacting, picking and growing for unforgetting continual learning,''
  \emph{Advances in Neural Information Processing Systems}, vol.~32, 2019.

\bibitem{method_exstream_hayes_2019}
T.~L. Hayes, N.~D. Cahill, and C.~Kanan, ``Memory efficient experience replay
  for streaming learning,'' \emph{2019 International Conference on Robotics and
  Automation (ICRA)}, 2019.

\bibitem{method_icarl_rebuffi_2017}
S.-A. Rebuffi, A.~Kolesnikov, G.~Sperl, and C.~H. Lampert, ``icarl: Incremental
  classifier and representation learning,'' \emph{2017 IEEE Conference on
  Computer Vision and Pattern Recognition (CVPR)}, 2017.

\bibitem{overviewSpatiojainStructuralrnnDeepLearning2016}
A.~Jain, A.~R. Zamir, S.~Savarese, and A.~Saxena, ``Structural-rnn: {{Deep}}
  learning on spatio-temporal graphs,'' in \emph{Proceedings of the Ieee
  Conference on Computer Vision and Pattern Recognition}, 2016, pp. 5308--5317.

\bibitem{method_stgcn_yu2017}
B.~Yu, H.~Yin, and Z.~Zhu, ``Spatio-temporal graph convolutional networks:
  {{A}} deep learning framework for traffic forecasting,'' \emph{arXiv preprint
  arXiv:1709.04875}, 2017.

\bibitem{overviewSpatiozhangGaanGatedAttention2018}
J.~Zhang, X.~Shi, J.~Xie, H.~Ma, I.~King, and D.-Y. Yeung, ``Gaan: {{Gated}}
  attention networks for learning on large and spatiotemporal graphs,''
  \emph{arXiv preprint arXiv:1803.07294}, 2018.

\bibitem{spatio_attention_guo2019}
S.~Guo, Y.~Lin, N.~Feng, C.~Song, and H.~Wan, ``Attention based
  spatial-temporal graph convolutional networks for traffic flow forecasting,''
  in \emph{Proceedings of the {{AAAI}} Conference on Artificial Intelligence},
  vol.~33, 2019, pp. 922--929.

\bibitem{method_FGN_wang_2020}
C.~Wang, D.~Gao, Y.~Qiu, and S.~Scherer, ``Lifelong graph learning,'' in
  \emph{2022 Conference on Computer Vision and Pattern Recognition (CVPR)},
  2022.

\bibitem{benchmark_carta2021catastrophic}
A.~Carta, A.~Cossu, F.~Errica, and D.~Bacciu, ``Catastrophic forgetting in deep
  graph networks: an introductory benchmark for graph classification,''
  \emph{arXiv preprint arXiv:2103.11750}, 2021.

\bibitem{zhuang2020comprehensiveTransfer}
F.~Zhuang, Z.~Qi, K.~Duan, D.~Xi, Y.~Zhu, H.~Zhu, H.~Xiong, and Q.~He, ``A
  comprehensive survey on transfer learning,'' \emph{Proceedings of the IEEE},
  vol. 109, no.~1, pp. 43--76, 2020.

\bibitem{zhang2021surveyMultitask}
Y.~Zhang and Q.~Yang, ``A survey on multi-task learning,'' \emph{IEEE
  Transactions on Knowledge and Data Engineering}, 2021.

\bibitem{hoi2021online}
S.~C. Hoi, D.~Sahoo, J.~Lu, and P.~Zhao, ``Online learning: A comprehensive
  survey,'' \emph{Neurocomputing}, vol. 459, pp. 249--289, 2021.

\bibitem{yang2021generalized_OOD}
J.~Yang, K.~Zhou, Y.~Li, and Z.~Liu, ``Generalized out-of-distribution
  detection: A survey,'' \emph{arXiv preprint arXiv:2110.11334}, 2021.

\bibitem{li2022out_graphood}
H.~Li, X.~Wang, Z.~Zhang, and W.~Zhu, ``Out-of-distribution generalization on
  graphs: A survey,'' \emph{arXiv preprint arXiv:2202.07987}, 2022.

\bibitem{lesort2021understanding_continual_OOD}
T.~Lesort, M.~Caccia, and I.~Rish, ``Understanding continual learning settings
  with data distribution drift analysis,'' \emph{arXiv preprint
  arXiv:2104.01678}, 2021.

\bibitem{yang2021discrete}
M.~Yang, M.~Zhou, M.~Kalander, Z.~Huang, and I.~King, ``Discrete-time temporal
  network embedding via implicit hierarchical learning in hyperbolic space,''
  in \emph{Proceedings of the 27th ACM SIGKDD Conference on Knowledge Discovery
  \& Data Mining}, 2021, pp. 1975--1985.

\bibitem{skarding2021foundations}
J.~Skarding, B.~Gabrys, and K.~Musial, ``Foundations and modeling of dynamic
  networks using dynamic graph neural networks: A survey,'' \emph{IEEE Access},
  vol.~9, pp. 79\,143--79\,168, 2021.

\bibitem{ma2020streaming}
Y.~Ma, Z.~Guo, Z.~Ren, J.~Tang, and D.~Yin, ``Streaming graph neural
  networks,'' in \emph{Proceedings of the 43rd International ACM SIGIR
  Conference on Research and Development in Information Retrieval}, 2020, pp.
  719--728.

\bibitem{method_HPNs_zhang_2021}
S.~Sun, Q.~Sun, K.~Zhou, and T.~Lv, ``Hierarchical attention prototypical
  networks for few-shot text classification,'' in \emph{Proceedings of the 2019
  conference on empirical methods in natural language processing and the 9th
  international joint conference on natural language processing
  (EMNLP-IJCNLP)}, 2019, pp. 476--485.

\bibitem{method_DiCGRL_kou_2020}
X.~Kou, Y.~Lin, S.~Liu, P.~Li, J.~Zhou, and Y.~Zhang, ``Disentangle-based
  {{Continual Graph Representation Learning}},'' \emph{arXiv:2010.02565 [cs]},
  Nov. 2020.

\bibitem{method_GFCIL_zheng_2022}
Z.~Tan, K.~Ding, R.~Guo, and H.~Liu, ``Graph few-shot class-incremental
  learning,'' in \emph{Proceedings of the Fifteenth ACM International
  Conference on Web Search and Data Mining}, 2022, pp. 987--996.

\bibitem{method_TWP_liu_2020}
H.~Liu, Y.~Yang, and X.~Wang, ``Overcoming catastrophic forgetting in graph
  neural networks,'' in \emph{Proceedings of the AAAI Conference on Artificial
  Intelligence}, vol.~35, no.~10, 2021, pp. 8653--8661.

\bibitem{method_translationBased_song_2018}
H.-J. Song and S.-B. Park, ``Enriching translation-based knowledge graph
  embeddings through continual learning,'' \emph{IEEE Access}, vol.~6, pp.
  60\,489--60\,497, 2018.

\bibitem{method_LDANE_wei_2019}
H.~Wei, G.~Hu, W.~Bai, S.~Xia, and Z.~Pan, ``Lifelong representation learning
  in dynamic attributed networks,'' \emph{Neurocomputing}, vol. 358, pp. 1--9,
  Sep. 2019.

\bibitem{chen2018fastgcn}
J.~Chen, T.~Ma, and C.~Xiao, ``Fastgcn: fast learning with graph convolutional
  networks via importance sampling,'' \emph{arXiv preprint arXiv:1801.10247},
  2018.

\bibitem{methods_FGNminibatch_kri_2017}
A.~Krizhevsky, I.~Sutskever, and G.~E. Hinton, ``Imagenet classification with
  deep convolutional neural networks,'' \emph{Communications of the ACM},
  vol.~60, no.~6, pp. 84--90, 2017.

\bibitem{bench_coraCiteseerPubmed_sen2008}
P.~Sen, G.~Namata, M.~Bilgic, L.~Getoor, B.~Galligher, and T.~Eliassi-Rad,
  ``Collective classification in network data,'' \emph{AI magazine}, vol.~29,
  no.~3, pp. 93--93, 2008.

\bibitem{bench_arxiv_mikolov2013distributed}
T.~Mikolov, I.~Sutskever, K.~Chen, G.~S. Corrado, and J.~Dean, ``Distributed
  representations of words and phrases and their compositionality,''
  \emph{Advances in neural information processing systems}, vol.~26, 2013.

\bibitem{bench_arxiv_wang2020microsoft}
K.~Wang, Z.~Shen, C.~Huang, C.-H. Wu, Y.~Dong, and A.~Kanakia, ``Microsoft
  academic graph: When experts are not enough,'' \emph{Quantitative Science
  Studies}, vol.~1, no.~1, pp. 396--413, 2020.

\bibitem{dicgrl_ne_GAN_velivckovic2017}
P.~Veli{\v{c}}kovi{\'c}, G.~Cucurull, A.~Casanova, A.~Romero, P.~Lio, and
  Y.~Bengio, ``Graph attention networks,'' \emph{arXiv preprint
  arXiv:1710.10903}, 2017.

\bibitem{klicpera2018predict_APPNP}
J.~Klicpera, A.~Bojchevski, and S.~G{\"u}nnemann, ``Predict then propagate:
  Graph neural networks meet personalized pagerank,'' \emph{arXiv preprint
  arXiv:1810.05997}, 2018.

\bibitem{xu2018powerful}
K.~Xu, W.~Hu, J.~Leskovec, and S.~Jegelka, ``How powerful are graph neural
  networks?'' \emph{arXiv preprint arXiv:1810.00826}, 2018.

\bibitem{pei2020geom_datasetActor}
H.~Pei, B.~Wei, K.~C.-C. Chang, Y.~Lei, and B.~Yang, ``Geom-gcn: Geometric
  graph convolutional networks,'' \emph{arXiv preprint arXiv:2002.05287}, 2020.

\bibitem{method_ergnnCoverage_tim_2016}
T.~de~Bruin, J.~Kober, K.~Tuyls, and R.~Babuška, ``Improved deep reinforcement
  learning for robotics through distribution-based experience retention,'' in
  \emph{2016 IEEE/RSJ International Conference on Intelligent Robots and
  Systems (IROS)}, 2016, pp. 3947--3952.

\bibitem{method_ergnncls_kumaran_2016}
D.~Kumaran, D.~Hassabis, and J.~L. McClelland, ``What learning systems do
  intelligent agents need? complementary learning systems theory updated,''
  \emph{Trends in Cognitive Sciences}, vol.~20, no.~7, pp. 512--534, 2016.

\bibitem{method_ergnnAdam_kingma_2015}
\BIBentryALTinterwordspacing
D.~P. Kingma and J.~Ba, ``Adam: {A} method for stochastic optimization,'' in
  \emph{3rd International Conference on Learning Representations, {ICLR} 2015,
  San Diego, CA, USA, May 7-9, 2015, Conference Track Proceedings}, Y.~Bengio
  and Y.~LeCun, Eds., 2015. [Online]. Available:
  \url{http://arxiv.org/abs/1412.6980}
\BIBentrySTDinterwordspacing

\bibitem{perozzi2014deepwalk}
B.~Perozzi, R.~Al-Rfou, and S.~Skiena, ``Deepwalk: Online learning of social
  representations,'' in \emph{Proceedings of the 20th ACM SIGKDD international
  conference on Knowledge discovery and data mining}, 2014, pp. 701--710.

\bibitem{grover2016node2vec}
A.~Grover and J.~Leskovec, ``node2vec: Scalable feature learning for
  networks,'' in \emph{Proceedings of the 22nd ACM SIGKDD international
  conference on Knowledge discovery and data mining}, 2016, pp. 855--864.

\bibitem{method_openworldIsotro_dwivedi_2020}
\BIBentryALTinterwordspacing
V.~P. Dwivedi, C.~K. Joshi, T.~Laurent, Y.~Bengio, and X.~Bresson,
  ``Benchmarking graph neural networks,'' \emph{CoRR}, vol. abs/2003.00982,
  2020. [Online]. Available: \url{https://arxiv.org/abs/2003.00982}
\BIBentrySTDinterwordspacing

\bibitem{tang2008arnetminer_dblbDataset}
J.~Tang, J.~Zhang, L.~Yao, J.~Li, L.~Zhang, and Z.~Su, ``Arnetminer: extraction
  and mining of academic social networks,'' in \emph{Proceedings of the 14th
  ACM SIGKDD international conference on Knowledge discovery and data mining},
  2008, pp. 990--998.

\bibitem{dicgrl_kgs_rotate_sun2019G}
Z.~Sun, Z.-H. Deng, J.-Y. Nie, and J.~Tang, ``Rotate: Knowledge graph embedding
  by relational rotation in complex space,'' \emph{arXiv preprint
  arXiv:1902.10197}, 2019.

\bibitem{dicgrl_kgs_transe_bordes2013}
A.~Bordes, N.~Usunier, A.~Garcia-Duran, J.~Weston, and O.~Yakhnenko,
  ``Translating embeddings for modeling multi-relational data,'' \emph{Advances
  in neural information processing systems}, vol.~26, 2013.

\bibitem{dicgrl_ne_DGCN_henaff2015}
M.~Henaff, J.~Bruna, and Y.~LeCun, ``Deep convolutional networks on
  graph-structured data,'' \emph{arXiv preprint arXiv:1506.05163}, 2015.

\bibitem{toutanova2015observed_FBdataset}
K.~Toutanova and D.~Chen, ``Observed versus latent features for knowledge base
  and text inference,'' in \emph{Proceedings of the 3rd workshop on continuous
  vector space models and their compositionality}, 2015, pp. 57--66.

\bibitem{dettmers2018convolutional_WNdataset}
T.~Dettmers, P.~Minervini, P.~Stenetorp, and S.~Riedel, ``Convolutional 2d
  knowledge graph embeddings,'' in \emph{Proceedings of the AAAI conference on
  artificial intelligence}, vol.~32, no.~1, 2018.

\bibitem{nguyen2017novel_convkb}
D.~Q. Nguyen, T.~D. Nguyen, D.~Q. Nguyen, and D.~Phung, ``A novel embedding
  model for knowledge base completion based on convolutional neural network,''
  \emph{arXiv preprint arXiv:1712.02121}, 2017.

\bibitem{gfcil_scaleFactor_lin2017}
T.-Y. Lin, P.~Goyal, R.~Girshick, K.~He, and P.~Doll{\'a}r, ``Focal loss for
  dense object detection,'' in \emph{Proceedings of the IEEE international
  conference on computer vision}, 2017, pp. 2980--2988.

\bibitem{gfcil_scaleFactor_menon2020l}
A.~K. Menon, S.~Jayasumana, A.~S. Rawat, H.~Jain, A.~Veit, and S.~Kumar,
  ``Long-tail learning via logit adjustment,'' \emph{arXiv preprint
  arXiv:2007.07314}, 2020.

\bibitem{mcauley2015inferring_amazon}
J.~McAuley, R.~Pandey, and J.~Leskovec, ``Inferring networks of substitutable
  and complementary products,'' in \emph{Proceedings of the 21th ACM SIGKDD
  international conference on knowledge discovery and data mining}, 2015, pp.
  785--794.

\bibitem{zhang2021few_CECmethod}
C.~Zhang, N.~Song, G.~Lin, Y.~Zheng, P.~Pan, and Y.~Xu, ``Few-shot incremental
  learning with continually evolved classifiers,'' in \emph{Proceedings of the
  IEEE/CVF Conference on Computer Vision and Pattern Recognition}, 2021, pp.
  12\,455--12\,464.

\bibitem{method_SI_zenke2017}
F.~Zenke, B.~Poole, and S.~Ganguli, ``Continual learning through synaptic
  intelligence,'' in \emph{International Conference on Machine Learning}.\hskip
  1em plus 0.5em minus 0.4em\relax PMLR, 2017, pp. 3987--3995.

\bibitem{bojchevski2017deep_corafulldataset}
A.~Bojchevski and S.~G{\"u}nnemann, ``Deep gaussian embedding of graphs:
  Unsupervised inductive learning via ranking,'' \emph{arXiv preprint
  arXiv:1707.03815}, 2017.

\bibitem{method_GDM_parisi2018}
G.~I. Parisi, J.~Tani, C.~Weber, and S.~Wermter, ``Lifelong learning of
  spatiotemporal representations with dual-memory recurrent
  self-organization,'' \emph{Frontiers in neurorobotics}, p.~78, 2018.

\bibitem{method_continualGNNReservoir}
J.~S. Vitter, ``Random sampling with a reservoir,'' \emph{ACM Transactions on
  Mathematical Software (TOMS)}, vol.~11, no.~1, pp. 37--57, 1985.

\bibitem{weber2019anti_ellipticDataset}
M.~Weber, G.~Domeniconi, J.~Chen, D.~K.~I. Weidele, C.~Bellei, T.~Robinson, and
  C.~E. Leiserson, ``Anti-money laundering in bitcoin: Experimenting with graph
  convolutional networks for financial forensics,'' \emph{arXiv preprint
  arXiv:1908.02591}, 2019.

\bibitem{benyahia2016dancer_syndataset}
O.~Benyahia, C.~Largeron, B.~Jeudy, and O.~R. Za{\"\i}ane, ``Dancer: Dynamic
  attributed network with community structure generator,'' in \emph{Joint
  European Conference on Machine Learning and Knowledge Discovery in
  Databases}.\hskip 1em plus 0.5em minus 0.4em\relax Springer, 2016, pp.
  41--44.

\bibitem{wang2016structural_SDNE}
D.~Wang, P.~Cui, and W.~Zhu, ``Structural deep network embedding,'' in
  \emph{Proceedings of the 22nd ACM SIGKDD international conference on
  Knowledge discovery and data mining}, 2016, pp. 1225--1234.

\bibitem{method_lifelongSentiment_chen_2015}
Z.~Chen, N.~Ma, and B.~Liu, ``Lifelong learning for sentiment classification,''
  2018.

\bibitem{chen2001freeway_trafficstreamDataset}
C.~Chen, K.~Petty, A.~Skabardonis, P.~Varaiya, and Z.~Jia, ``Freeway
  performance measurement system: mining loop detector data,''
  \emph{Transportation Research Record}, vol. 1748, no.~1, pp. 96--102, 2001.

\bibitem{song2020spatial_stsgcn}
C.~Song, Y.~Lin, S.~Guo, and H.~Wan, ``Spatial-temporal synchronous graph
  convolutional networks: A new framework for spatial-temporal network data
  forecasting,'' in \emph{Proceedings of the AAAI Conference on Artificial
  Intelligence}, vol.~34, no.~01, 2020, pp. 914--921.

\bibitem{cho2014learning_GRU}
K.~Cho, B.~Van~Merri{\"e}nboer, C.~Gulcehre, D.~Bahdanau, F.~Bougares,
  H.~Schwenk, and Y.~Bengio, ``Learning phrase representations using rnn
  encoder-decoder for statistical machine translation,'' \emph{arXiv preprint
  arXiv:1406.1078}, 2014.

\bibitem{qu2020generalOptimize}
R.~Qu, G.~Kendall, and N.~Pillay, ``The general combinatorial optimization
  problem: Towards automated algorithm design,'' \emph{IEEE Computational
  Intelligence Magazine}, vol.~15, no.~2, pp. 14--23, 2020.

\bibitem{bench_products_chiang2019cluster}
W.-L. Chiang, X.~Liu, S.~Si, Y.~Li, S.~Bengio, and C.-J. Hsieh, ``Cluster-gcn:
  An efficient algorithm for training deep and large graph convolutional
  networks,'' in \emph{Proceedings of the 25th ACM SIGKDD International
  Conference on Knowledge Discovery \& Data Mining}, 2019, pp. 257--266.

\bibitem{bench_proteins_szklarczyk2019string}
D.~Szklarczyk, A.~L. Gable, D.~Lyon, A.~Junge, S.~Wyder, J.~Huerta-Cepas,
  M.~Simonovic, N.~T. Doncheva, J.~H. Morris, P.~Bork \emph{et~al.}, ``String
  v11: protein--protein association networks with increased coverage,
  supporting functional discovery in genome-wide experimental datasets,''
  \emph{Nucleic acids research}, vol.~47, no.~D1, pp. D607--D613, 2019.

\bibitem{metrics_second_chaudhry2018efficient}
A.~Chaudhry, M.~Ranzato, M.~Rohrbach, and M.~Elhoseiny, ``Efficient lifelong
  learning with a-gem,'' \emph{arXiv preprint arXiv:1812.00420}, 2018.

\bibitem{openIssue_catastrophicInterference_gordon}
M.~McCloskey and N.~J. Cohen, ``Catastrophic interference in connectionist
  networks: The sequential learning problem,'' in \emph{Psychology of learning
  and motivation}.\hskip 1em plus 0.5em minus 0.4em\relax Elsevier, 1989,
  vol.~24, pp. 109--165.

\bibitem{min2022transformer_graphTrans}
E.~Min, R.~Chen, Y.~Bian, T.~Xu, K.~Zhao, W.~Huang, P.~Zhao, J.~Huang,
  S.~Ananiadou, and Y.~Rong, ``Transformer for graphs: An overview from
  architecture perspective,'' \emph{arXiv preprint arXiv:2202.08455}, 2022.

\bibitem{zhang2021surveyFederated}
C.~Zhang, Y.~Xie, H.~Bai, B.~Yu, W.~Li, and Y.~Gao, ``A survey on federated
  learning,'' \emph{Knowledge-Based Systems}, vol. 216, p. 106775, 2021.

\bibitem{zhang2021federated}
H.~Zhang, T.~Shen, F.~Wu, M.~Yin, H.~Yang, and C.~Wu, ``Federated graph
  learning--a position paper,'' \emph{arXiv preprint arXiv:2105.11099}, 2021.

\bibitem{usmanova2021distillation}
A.~Usmanova, F.~Portet, P.~Lalanda, and G.~Vega, ``A distillation-based
  approach integrating continual learning and federated learning for pervasive
  services,'' \emph{arXiv preprint arXiv:2109.04197}, 2021.

\end{thebibliography}

\end{document}